\PassOptionsToPackage{table}{xcolor}
\documentclass{article} 
\usepackage[preprint]{colm2026_conference}

\usepackage[T1]{fontenc}
\usepackage{microtype}
\usepackage{hyperref}
\usepackage{url}

\usepackage{amsmath}
\usepackage{amssymb}
\usepackage{mathtools}
\usepackage{amsthm}
\newtheorem{example}{Example}

\usepackage{booktabs}
\usepackage{multirow}
\usepackage{array}

\usepackage{xcolor}
\usepackage{colortbl}

\usepackage{tikz}
\usetikzlibrary{arrows.meta, positioning, shapes.geometric, fit, backgrounds, calc, automata}
\usepackage{algorithm}
\usepackage{algpseudocode}
\usepackage{placeins}
\usepackage{pifont}
\usepackage{enumitem}

\usepackage{listings}
\lstdefinestyle{codestyle}{
  basicstyle=\ttfamily\scriptsize,
  breaklines=true,
  frame=single,
  backgroundcolor=\color{black!4},
  rulecolor=\color{black!20},
  escapeinside={(*@}{@*)},
}
\usepackage{makecell}

\usepackage{lineno}
\usepackage[font=small,labelfont=bf]{caption}
\usepackage[textsize=tiny]{todonotes}

\usepackage[normalem]{ulem}

\definecolor{aiblue}{RGB}{0,50,160}

\definecolor{hardgray}{gray}{0.90}
\definecolor{dgreen}{rgb}{0.33, 0.66, 0.49}    
\definecolor{mgreen}{rgb}{0.5, 0.78, 0.62}    
\definecolor{lgreen}{rgb}{0.78, 0.91, 0.82}   
\definecolor{lred}{rgb}{0.98, 0.9, 0.88}      
\definecolor{mred}{rgb}{0.95, 0.7, 0.65}      
\definecolor{dred}{rgb}{0.85, 0.45, 0.4}      
\definecolor{darkblue}{rgb}{0, 0, 0.5}

\newcommand{\dgcell}[1]{\cellcolor{dgreen}#1}
\newcommand{\mgcell}[1]{\cellcolor{mgreen}#1}
\newcommand{\lgcell}[1]{\cellcolor{lgreen}#1}
\newcommand{\lrcell}[1]{\cellcolor{lred}#1}
\newcommand{\mrcell}[1]{\cellcolor{mred}#1}
\newcommand{\drcell}[1]{\cellcolor{dred}#1}
\newcommand{\graycell}[1]{\cellcolor{hardgray}#1}

\hypersetup{colorlinks=true, citecolor=darkblue, linkcolor=darkblue, urlcolor=darkblue}

\title{\textsc{TempoBench}: Evaluating Temporal Causal Reasoning in Large Language Models}

\author{Nikolaus Holzer \\
Columbia University \\
\texttt{holzer@cs.columbia.edu} \\
\And
William Fishell \\
Columbia University \\
\texttt{wf2322@columbia.edu} \\
\And
Baishakhi Ray \\
Columbia University \\
\texttt{rayb@cs.columbia.edu} \\
\And
Mark Santolucito \\
Columbia University, Barnard College \\
\texttt{msantolu@barnard.edu}
}

\begin{document}

\ifcolmsubmission
\linenumbers
\fi

\maketitle

\begin{abstract}
Temporal reasoning involves understanding how systems evolve over time through input-driven state transitions. A key aspect is temporal causal reasoning, causally reasoning about what prior inputs were necessary in causing an observed outcome. While large language models (LLMs) perform well at forward simulation, predicting outputs from inputs, they struggle to identify the minimal causal inputs of outcomes. To study this distinction, we define two tasks: \textit{trace simulation} (SIM), which requires models to simulate system execution, and \textit{minimal causal attribution} (MIN), which identifies the minimal set of inputs necessary for a given outcome. We introduce \textsc{TempoBench}, the first formally verified benchmark for temporal causal reasoning, built from synthesized Mealy machines with controllable complexity and provably correct causal labels. Across frontier models, we observe that despite achieving up to 96\% accuracy on the SIM task, performance on the causal attribution MIN task drops below 25\%; models fail to reason about causal necessity. Over 94\% of causal errors involve overspecification, where models perform retrieval and list all possible inputs rather than reasoning about the minimal causal subset. Fine-tuning on \textsc{TempoBench} training corpus improves causal reasoning and generalizes better than math, code, or instruction training, with gains across standard reasoning benchmarks.
\end{abstract}

\section{Introduction}
\label{sec:intro}

Temporal reasoning is a fundamental capability for intelligent systems operating in dynamic environments. At its core, it involves understanding how a system evolves over time as inputs drive transitions between states and produce observable outcomes. This capability underlies a wide range of real-world settings, including program execution, control systems, decision-making agents, and interactive workflows, where reasoning requires not just static inference but tracking how actions unfold over time. In such settings, success depends on the ability to reason over execution traces that capture sequences of inputs, intermediate states, and outputs that encode system behavior.

However, effective temporal reasoning extends beyond forward simulation: in many practical scenarios, the goal is not only to predict what will happen, but to explain why something happened. This requires identifying which prior inputs were necessary for an observed outcome, i.e., reasoning about causal dependence within a temporal process. While large language models (LLMs) have approached or surpassed human-level performance on benchmarks in mathematical reasoning~\cite{cobbe2021trainingverifierssolvemath}, coding~\cite{jimenez2024swebench}, and problem-solving~\cite{bubeck2023sparksartificialgeneralintelligence, chen2023program, wang2023selfconsistency, 10.5555/3666122.3666639ToT, 10.1609/aaai.v38i16.29720GOT}, they are predominantly trained on data that encourages forward reasoning via simulated executions. The common \textit{"if-this-then-that"} chain-of-thought (CoT) paradigm reinforces a forward causal view, enabling strong performance on trace simulation but offering limited support for reasoning backwards about observed outcomes.

This limitation becomes especially pronounced in agentic settings, where decisions unfold over long horizons, and errors compound over time, making retrospective debugging and diagnosis critical. In such contexts, the ability to perform causal attribution over execution traces, identifying the minimal inputs responsible for an outcome, is essential. However, it remains unclear whether current LLMs can reliably perform this form of backward, causally grounded reasoning, motivating the need for both targeted evaluation and training frameworks for temporal causal reasoning~\cite{ikram2022root}.

\textbf{Key Challenge.} A bottleneck in advancing temporal reasoning is the lack of reliable training data. Existing sources are either fully verifiable but limited in scope (e.g., mathematical proofs) or realistic but noisy and weakly supervised (e.g., code, execution traces, or distillation data). While code can be tested, these signals are often incomplete, expensive to obtain, and do not capture minimal causal dependencies. As a result, current datasets fail to provide the combination of flexibility, scale, and strong correctness guarantees needed to train models for temporal causal reasoning.

\textbf{Proposed Solution.} To overcome this, we propose using automata as a source of infinitely scalable, fully verifiable reasoning data. At a high level, an automaton is a simple model of a system that evolves step-by-step: at each time step, it receives a set of inputs and deterministically produces an output and a new state. Figure~\ref{fig:pipeline} illustrates this with a code review bot. Each day, the system observes signals such as whether tests pass and whether a reviewer approves, and based on these inputs, transitions between states like \textit{idle}, \textit{review}, or \textit{merged}, while producing outputs such as \textit{proceed} or \textit{hold}. By unrolling this process over time, we obtain execution traces that resemble real workflows. A key advantage of using automata is that, unlike real-world data, the underlying rules of the system are fully known. This allows us not only to simulate what happens at each step, but also to precisely determine which inputs were actually necessary for a given outcome. For example, in Figure~\ref{fig:pipeline}, a pull request may be put on hold whenever there is no approval, regardless of test results, revealing that a lack of approval is the true causal factor for the delayed pull request. This ability to generate diverse traces with exact, minimal causal explanations enables scalable and reliable supervision for temporal causal reasoning.

\begin{figure*}[t]
    \centering
    \resizebox{\textwidth}{!}{%
\begin{tikzpicture}[scale=0.17]
\tikzstyle{every node}+=[inner sep=0pt]


\fill[black!4, rounded corners=6pt] (8,-56) rectangle (73,-10);
\draw[black!15, rounded corners=6pt, line width=0.4pt] (8,-56) rectangle (73,-10);

\draw (40.5,-12.5) node {\large\bfseries Code Review Bot};

\draw (10,-16) node[anchor=north west, text width=300] {\footnotesize
A CI/CD bot manages pull requests. Each round, two signals arrive: whether \textbf{tests pass} (T) and whether a reviewer \textbf{approves} (A).};

\draw (10,-23) node[anchor=north west, text width=300] {\footnotesize
Through observation of the system the following patterns have surfaced: If both signals are positive, the PR \textit{proceeds} to review. If tests pass but there is no approval, it gets \textit{flagged}. If neither signal is positive, it is \textit{escalated}. From review, the only route forward is to merge.};

\draw[black!20, line width=0.4pt] (10,-33) -- (71,-33);

\draw (10,-34.5) node[anchor=north west] {\footnotesize\bfseries Observed trace:};
\draw (10,-37) node[anchor=north west, text width=300] {\footnotesize
\textbf{Day 1:} Tests pass, reviewer approves $\to$ PR \textit{proceeds}. \enspace (Idle $\to$ Rev)};
\draw (10,-40.5) node[anchor=north west, text width=300] {\footnotesize
\textbf{Day 2:} Tests fail, no approval $\to$ PR put on \textit{hold}. \enspace (Rev $\to$ Merged)};

\draw[black!20, line width=0.4pt] (10,-44) -- (71,-44);

\draw (10,-45.5) node[anchor=north west] {\footnotesize\bfseries Which inputs actually mattered?};
\draw (10,-48) node[anchor=north west, text width=300] {\footnotesize
\textcolor{dgreen}{\textbf{Day 1:}} Both needed. Without tests $\to$ hold. Without approval $\to$ hold.};
\draw (10,-51.5) node[anchor=north west, text width=300] {\footnotesize
\textcolor{dgreen}{\textbf{Day 2:}} No constraints. Only one transition from Rev; any input gives \textit{hold}.};

\draw[->, line width=1.2pt, black!35] (74,-32) -- (76,-32);


\fill[black!4, rounded corners=6pt] (77,-56) rectangle (125,-10);
\draw[black!12, rounded corners=6pt, line width=0.4pt] (77,-56) rectangle (125,-10);

\draw (101,-12.5) node {\large\bfseries Approximated Automata};

\draw [black] (91.7,-21.7) circle (3);
\draw (91.7,-21.7) node {\footnotesize Idle};
\draw [black] (105.8,-21.7) circle (3);
\draw (105.8,-21.7) node {\footnotesize Rev};
\draw [black] (91.7,-35.1) circle (3);
\draw (91.7,-35.1) node {\footnotesize Flag};
\draw [black] (105.8,-35.1) circle (3);
\draw (105.8,-35.1) node {\footnotesize Esc};
\draw [black] (117.9,-28.4) circle (3);
\draw (117.9,-28.4) node {\footnotesize Mrgd};

\draw [black] (83.9,-21.7) -- (88.7,-21.7);
\fill [black] (88.7,-21.7) -- (87.9,-21.2) -- (87.9,-22.2);

\draw [black] (94.7,-21.7) -- (102.8,-21.7);
\fill [black] (102.8,-21.7) -- (102,-21.2) -- (102,-22.2);
\draw (98.75,-20.5) node [above] {\footnotesize $T,A/p$};

\draw [black] (91.7,-24.7) -- (91.7,-32.1);
\fill [black] (91.7,-32.1) -- (92.2,-31.3) -- (91.2,-31.3);
\draw (90,-28.4) node [left] {\footnotesize $T,{!}A/h$};

\draw [black] (93.87,-23.77) -- (103.63,-33.03);
\fill [black] (103.63,-33.03) -- (103.39,-32.12) -- (102.7,-32.84);
\draw (95.59,-29.2) node [below] {\footnotesize ${!}T,{!}A/h$};

\draw [black] (108.92,-33.65) -- (115.28,-29.85);
\fill [black] (115.28,-29.85) -- (114.33,-29.8) -- (114.82,-30.68);
\draw (115,-32.5) node [below] {\footnotesize ${!}T,A/p$};

\draw [black] (108.92,-23.15) -- (115.28,-26.95);
\fill [black] (115.28,-26.95) -- (114.82,-26.12) -- (114.33,-27);
\draw (115.2,-24.2) node [above] {\footnotesize ${!}T,{!}A/h$};

\draw [black] (89.02,-36.423) arc (324:36:2.25);
\draw (84.2,-35.1) node [left] {\footnotesize ${!}T,{!}A/h$};
\fill [black] (89.02,-33.78) -- (88.67,-32.9) -- (88.08,-33.71);

\draw (101,-43) node {\footnotesize\itshape T\,=\,tests\_pass,\; A\,=\,approved};
\draw (101,-45.5) node {\footnotesize\itshape p\,=\,proceed,\; h\,=\,hold};

\draw (101,-50) node {\normalsize\itshape Trace: Idle $\xrightarrow{T,A}$ Rev $\xrightarrow{{!}T,{!}A}$ Mrgd};

\end{tikzpicture}%
    }
    \vspace{-8pt}
    \caption{A natural language description of a code review bot and a corresponding automata representation of the system's rules. The machine has two boolean inputs ($T$\,=\,tests\_pass, $A$\,=\,approved) and two outputs ($p$\,=\,proceed, $h$\,=\,hold). SIM requires tracing the execution; MIN requires identifying which inputs were causally necessary for each output.}
    \label{fig:pipeline}
    \vspace{-6pt}
\end{figure*}

We present \textsc{TempoBench}, an infinitely scalable and fully verifiable dataset and benchmark for temporal causal reasoning, built on formally specified reactive systems. We synthesize automata from domains such as turn-based games and program-like controllers, and simulate their executions to generate traces of inputs, states, and outputs. Because the underlying system is fully known, we can compute minimal causal dependencies for each outcome and use them as supervision, yielding high-quality training data that supports both forward simulation and causal reasoning.

Using this framework, we find that while frontier reasoning models achieve up to 96\% accuracy on forward simulation, their performance drops below 25\% when asked to identify causal dependencies, revealing a fundamental gap in temporal reasoning. In contrast, models fine-tuned on \textsc{TempoBench} not only learn to perform causal attribution, but do so efficiently: even relatively small models (7B–8B) substantially close the gap and in some cases match or exceed much larger frontier models. This suggests that temporal causal reasoning is not simply an emergent property of scale, but a learnable capability enabled by the right training signal.

Importantly, fine-tuning on \textsc{TempoBench} preserves—and often improves—performance on standard benchmarks (e.g., GSM8K~\cite{cobbe2021trainingverifierssolvemath}, MMLU, ARC, BBH, HellaSwag, IFEval). At the 8B scale, mixed training with \textsc{TempoBench} yields the strongest overall models, improving performance by $+5$ to $+7$ points on average across benchmarks, while avoiding overfitting in domain-specific datasets, as is observed, for example, in OMI-1. These results indicate that the counterfactual reasoning structure underlying \textsc{TempoBench} transfers beyond temporal tasks to broader domains, including mathematical reasoning, reading comprehension, and knowledge recall. To our knowledge, this is the first benchmark to provide fully verifiable supervision for temporal causal reasoning with controllable structural complexity, enabling systematic study of the factors that make such reasoning difficult for current models.

Our key contributions are: (1)~\textsc{TempoBench}, the first formally verified benchmark for temporal causal reasoning, built on automata to generate infinitely scalable, fully verifiable reasoning data with controllable structural complexity and provably correct ground truth; (2)~a systematic evaluation of frontier models revealing a universal overspecification failure mode (over 94\% of causal errors), highlighting a fundamental gap in causal reasoning; and (3)~a training study across three model families showing that automata-generated data provides a more effective and balanced fine-tuning signal than math-only, code-only, or instruction data.

Overall, our work demonstrates that automata offer a principled way to approximate real world, noisy environments while retaining full verifiability, enabling the scalable generation of high quality supervision for temporal causal reasoning.

\section{Related Work}
\label{sec:related}
\textbf{LLM Reasoning Systems.} Recent advances in LLM-powered reasoning agents have driven research on benchmarking and LLM reasoning capabilities. Methods such as Chain of Thought (CoT)~\cite{wei2022chain}, self-consistency~\cite{wang2023selfconsistency}, and tree or graph-structured reasoning~\cite{10.5555/3666122.3666639ToT,10.1609/aaai.v38i16.29720GOT} improve accuracy on reasoning benchmarks by leveraging intermediate steps or tool use~\cite{chen2023program}. However, benchmarks that test reasoning find themselves either too rigid or unverifiable; for example, mathematical problem reasoning \cite{cobbe2021trainingverifierssolvemath}, coding benchmarks \cite{chen2021evaluatinglargelanguagemodels,  jimenez2024swebench}, challenging human-designed logic puzzles \cite{chollet2025arcagi2newchallengefrontier, chollet2025arcprize2024technical, srivastava2023beyond}, or generated quasi-formal benchmarks that incorporate verifiable structures \cite{fatemi2025test}. In contrast, \textsc{TempoBench} grounds reasoning evaluation in formal, parameterized systems, enabling quantitative study of how reasoning performance scales with structural problem features.

\textbf{Causality and Counterfactual Reasoning.} The \textsc{TempoBench} tasks are an instance of \textit{actual causality} in the sense of Halpern and Pearl (HP)~\cite{halpern2005causes}: given a fully observed trace, identify the minimal set of input assignments whose counterfactual alteration would change the outcome. This places \textsc{TempoBench} at the third (counterfactual) level of Pearl's causal hierarchy~\cite{pearl2009causality}. Unlike structural causal models (SCMs), which operate over DAGs with latent confounders, our setting uses deterministic Mealy machines where the causal structure is fully specified and verifiable; the CORP tool~\cite{finkbeiner2024synthesis} implements a variant of HP actual causality for reactive systems. Halpern's monograph~\cite{halpern2016actual} provides the broader theoretical framework. We note that the ``overspecification'' failure mode we identify, where models list sufficient but non-minimal causes, corresponds precisely to reporting a non-minimal witness in the HP definition.

\textbf{Temporal Reasoning Benchmarks.} Existing temporal reasoning benchmarks~\cite{fatemi2025test,wang2024tram, xiong2024large, tan-etal-2023-towards, JMLR:v25:22-1258} share four limitations: LLM-generated knowledge graphs risk training data contamination; ground-truth causal relationships are asserted rather than formally verified; fixed real-world topologies preclude systematic difficulty control; and tasks measure factual retrieval rather than counterfactual causal attribution. TestofTime (ToT)~\cite{fatemi2025test} partially addresses the third by using randomly generated graphs with tunable parameters, but provides no verified causal labels and does not isolate overspecification as a failure mode. \textsc{TempoBench} addresses all four: controllers are synthesised from formal specifications, causal labels are verified by CORP, structural complexity is fully parameterised, and the \textsc{TempoBench} tasks explicitly target counterfactual attribution.

\section{Formal Temporal Causal Reasoning}
\label{sec:formalism}
Rather than stochastic models like Markov decision processes, we focus on deterministic reactive systems to answer the question: given this sequence of inputs, how does the world respond? Our approach uses Mealy machine automata to model temporal reasoning. Using automata to model reasoning is well established in the ML community~\cite{wei2025satbench}. We explicitly model reasoning as agent interaction with a reactive world. 

A Mealy machine $M$ consists of a set of states $S$ and a start state $s_0$, transitions between the states $\delta$, and outputs $\lambda$ corresponding to each transition, as well as an alphabet of input $\Sigma$ and output $\Gamma$ symbols. 
\begin{equation}
    M=(S,s_0,\Sigma,\Gamma,\delta,\lambda)
\end{equation}
The transition and output functions are deterministic and defined as
\begin{equation}
\delta : S \times \Sigma \rightarrow S,
\qquad
\lambda : S \times \Sigma \rightarrow \Gamma .
\end{equation}
We assume the input alphabet factorizes as a tuple for $k$ boolean signals
\begin{equation}
\Sigma = \Sigma_1 \times \cdots \times \Sigma_k ,
\end{equation}
so that each transition consumes a vector-valued input
\begin{equation}
x_t = (x_t^{(1)}, \dots , x_t^{(k)}) \in \Sigma
\end{equation}
Given an input sequence $x_{1:n} = (x_1,\dots,x_n)$ with $x_t \in \Sigma$,
the induced execution of $M$ produces a state sequence $s_{0:n}$ and
output sequence $y_{1:n}$ defined recursively by
\begin{equation}
s_0 \text{ is the start state },
\ s_t = \delta(s_{t-1}, x_t),
\ y_t = \lambda(s_{t-1}, x_t),
\quad t = 1,\dots,n .
\label{eqn:recurrent-trace}
\end{equation}

A trace of length $n$ through a Mealy machine is a series of inputs and outputs that encode an execution
\begin{equation}
\pi = \bigl( (x_1,y_1), (x_2,y_2), \dots, (x_n,y_n) \bigr)
\in (\Sigma \times \Gamma)^n
\end{equation}
generated by unrolling equation~\ref{eqn:recurrent-trace}.

To define causality, fix an execution $x_{1:t}$ with output $y_t$. 
We define the causal set
$C \subseteq \{1,\dots,t\} \times \{1,\dots,k\}$ 
as the time-indexed input dimensions whose values are required for producing $y_t$, in the sense that only by changing the coordinates in $C$ can we alter the output $y_t$.
Formally, for all alternative traces $x'_{1:t}$, as long as $x_{1:t}$ and $x'_{1:t}$ agree on C, the output is the same:
\begin{equation}
\forall x'_{1:t} \quad, \
\left(
\forall (t',i)\in C,\; x_{t'}^{(i)} = x_{t'}'^{(i)}
\right)
\;\Longrightarrow\;
y_t = y_t'
\end{equation}
where $y_t'$ is the output at time $t$ induced by $x'_{1:t}$ under Equation~\ref{eqn:recurrent-trace}.
A causal set is minimal if no strict subset of $C$ satisfies this property. In a deterministic Mealy machine, the minimal causal set is equivalently the set of input dimensions whose values, if counterfactually changed, would alter the output; minimality ensures every element is necessary in the interventionist sense. Minimal causal set identification is conceptually the task of root cause identification in a reactive system~\cite{ikram2022root}.

The right panel of figure~\ref{fig:pipeline} shows a code review bot as a 5-state Mealy machine with inputs $T$ (tests\_pass) and $A$ (approved), and outputs $p$ (proceed) and $h$ (hold). For a trace of length 1: $\pi_1 = (T,{!}A) \to h$, the output $h$ occurs regardless of whether tests pass, as long as the PR is not approved. The minimal causal set is therefore $(*,{!}A)$. For a trace of length 2: $\pi_2 = ({!}T,{!}A)/h;\; ({!}T,A)/p$, examining the minimal causal set of $p$ at step~1 requires that the machine reaches Escalate, which depends on step~0's inputs. In this case, the minimal causal set found for $\pi_1$ no longer holds. This illustrates that the minimal causal set is not additive and must be recomputed for each observed output, because the state reached at each step depends on the full trace prefix.

We extract minimal causal sets by iterating over each step in the trace and checking which input constraints in the initial state are necessary to enforce the observed output under counterfactual intervention (see Algorithm~\ref{alg:counterfactual_causal_attribution} in Appendix~\ref{temporal causality}). The non-additivity of the minimal causal set (illustrated above) makes causal reasoning qualitatively harder than trace execution, as models cannot reduce it to local per-step retrieval.

\section{The \textsc{TempoBench} Dataset and Benchmark}
\label{sec:benchmark}
\textbf{\textsc{TempoBench}.}
is a benchmark for assessing an LLM's ability to perform key temporal reasoning tasks. We use reactive systems specified by temporal logics such as linear temporal logic (LTL) and synthesized using reactive synthesis~\cite{pnueli1989synthesis} to examine various system behaviors and examine the temporal properties of complex systems (See Appendix~\ref{LTL} for more information on reactive synthesis and LTL). To rigorously assess temporal reasoning, \textsc{TempoBench} focuses on two core tasks: \textbf{SIM (Trace Simulation)}, which measures a model's ability to simulate state transitions through a Mealy machine, and \textbf{MIN (Causal Minimality)}, which tests its ability to identify the minimal causal constraints required for observed outputs.

\textbf{Data Generation Pipeline.}
\textsc{TempoBench} is built on an end-to-end pipeline for generating data for SIM and MIN (Figure~\ref{fig:pipeline}). We use the SYNTCOMP benchmark~\cite{syntcomp_benchmarks}, a large standardized collection of reactive systems specified in temporal logic. Each system is expressed in LTL using the TLSF format~\cite{DBLP:journals/corr/Jacobs016}, which clearly distinguishes inputs, outputs, and their temporal behavior. Samples take on the order of seconds to generate using synthesis~\cite{renkin2022dissecting} and trace generation tools~\cite{10.1007/978-3-032-05435-7_3} on a single CPU core (see Appendix~\ref{app:synthesis}). Since we generate from verified structures, \textsc{TempoBench} requires no rejection sampling and no post generation verification. 

\textbf{Task 1: Trace Simulation (SIM).} Given a Mealy machine $M = (S, s_0, \Sigma, \Gamma, \delta, \lambda)$ and a finite input sequence $x_{1:n}$, produce the complete execution trace. For each time step $t$, the model must output the input read, output produced, and resulting state. This task unifies elements of runtime verification~\cite{leucker2009brief} and world modeling~\cite{ding2025understanding}. The model must reason over the system's transition dynamics to interpret the current state and navigate subsequent transitions. SIM measures how well models perform retrieval from structured inputs, as they must accurately parse and apply the automaton's state transitions.

\textbf{Task 2: Causal Minimality (MIN).}
Given a Mealy machine $M$, a trace $\pi$, and a target output $y_t$ observed at time step $t$, identify the minimal causal set of input constraints at each step $\tau \leq t$ that were necessary for $y_t$ to occur. At its core, this task tests the model's ability to reason causally through time---determining not just \emph{what} happened, but \emph{which inputs were essential} for the observed output. Examples of HOA representations are provided in Appendix~\ref{app:prompts}.

\section{Experiments}
\label{sec:experiments}
\begin{table*}[t]
  \centering
  \setlength{\tabcolsep}{2pt}
  \small
  \renewcommand{\arraystretch}{0.93}
  \resizebox{\textwidth}{!}{%
  \begin{tabular}{l|cc|cccc|cccc|cccc}
    \toprule
    & \multicolumn{2}{c|}{\textbf{SIM}}
    & \multicolumn{4}{c|}{\textbf{MIN}}
    & \multicolumn{4}{c|}{\textbf{MIN+}}
    & \multicolumn{4}{c}{\textbf{MIN-Hard}} \\
    \textbf{Model}
    & \makecell{Step} & \makecell{EM}
    & \makecell{Step} & \makecell{EM} & \makecell{Ovsp $\downarrow$} & \makecell{Unsp $\downarrow$}
    & \makecell{Step} & \makecell{EM} & \makecell{Ovsp $\downarrow$} & \makecell{Unsp $\downarrow$}
    & \makecell{Step} & \makecell{EM} & \makecell{Ovsp $\downarrow$} & \makecell{Unsp $\downarrow$} \\
    \midrule
    \multicolumn{15}{l}{\textit{Frontier models}} \\
    \cmidrule(l){1-15}
    Claude Sonnet 4.6  & 72.6 & 70.0 & 33.7 & 19.0 & 60.1 & 14.1 & 13.7 & 11.3 & 67.8 & 18.5 & 2.7 & 0.0 & 81.1 & 84.4 \\
    Claude Haiku 4.5   & 28.0 & 3.6 & 0.1 & 0.0 & 95.8 & 2.3 & 24.4 & 6.1 & 73.9 & 17.1 & 0.2 & 0.0 & 74.8 & 23.4 \\
    DeepSeek V3.2 (normal)& 23.5 & 3.0 & 0.1 & 0.0 & 96.2 & 2.3 & 24.2 & 4.2 & 75.2 & 17.1 & 19.5 & 0.0 & 75.9 & 23.8 \\
    DeepSeek V3        & 36.8 & 28.0 & 9.2 & 4.6 & 76.8 & 19.0 & 1.5 & 0.0 & 47.1 & 51.3 & 0.9 & 0.0 & 85.0 & 96.8 \\
    \cmidrule(l){1-15}
    \multicolumn{15}{l}{\textit{Reasoning models} (>20k tokens per sample)} \\
    \cmidrule(l){1-15}
    DeepSeek R1-0528   & 96.0 & 29.4 & 63.1 & 33.2 & 27.6 & 5.1 & 10.3 & 1.9 & 48.2 & 38.8 & 59.0 & 30.4 & 32.1 & 5.2 \\
    DeepSeek V3.2 (reason) & 77.8 & 51.7 & 55.2 & 32.9 & 14.9 & 5.0 & 9.7 & 0.5 & 30.9 & 37.8 & 53.3 & 31.2 & 16.2 & 4.8 \\
    \midrule
    \multicolumn{15}{l}{\textit{Best TB-trained per family}} \\
    \cmidrule(l){1-15}
    LLaMA-3.1-8B tb+SO  & 64.7 & 42.1 & 25.1 & 8.2 & 66.8 & 10.1 & 21.7 & 10.4 & 50.6 & 40.8 & 11.9 & 0.0 & 38.4 & 85.2 \\ 
    Qwen2.5-7B tb+OMI-1    & 75.0 & 50.3 & 22.4 & 7.6 & 67.6 & 13.9 & 10.3 & 9.5 & 57.1 & 44.5 & 0.1 & 0.0 & 27.8 & 99.8 \\ 
    LLaMA-3.2-3B tb+OC  & 3.8 & 1.8 & 12.2 & 3.8 & 81.4 & 15.4 & 10.3 & 7.0 & 71.8 & 58.5 & 0.3 & 0.0 & 35.0 & 93.7 \\ 
    \bottomrule
  \end{tabular}}
  \renewcommand{\arraystretch}{1.0}
  \caption{\textbf{\textsc{TempoBench} Evaluation: Frontier vs.\ Fine-tuned} (sample-level: each sample's step accuracy averaged across samples; unparsed samples scored as 0).
  \textit{MIN}: all causal steps.
  \textit{MIN+}: constrained steps only.
  \textit{MIN-Hard}: separate NC-free evaluation set.
  \textit{Ovsp}/\textit{Unsp}: over/underspecification.
  }
  \label{tab:llm_comp}
\end{table*}

\textbf{Model Evaluation.} We evaluate six frontier models: Claude Sonnet 4.6, Claude Haiku 4.5, DeepSeek V3, DeepSeek V3.2, DeepSeek R1, V3.2 with reasoning (reasoning results in Table~\ref{tab:retrieval}). We evaluate with 1500 samples, balanced across automaton types (Section~\ref{sec:benchmark}). Our evaluation prompt provides models with a natural-language view of the Mealy machine, all transition functions, the trace being processed, and the exact counterfactual algorithm to execute (Appendix~\ref{app:prompts}). We fine-tune LLaMA 3.1 8B~\cite{grattafiori_llama_2024}, LLaMA 3.2 3B~\cite{grattafiori_llama_2024}, and Qwen2.5-7B~\cite{qwen2025qwen25technicalreport} using LoRA~\cite{hu2022lora} with rank $r=64$, $\alpha=128$, and dropout 0.05, targeting the attention projection layers (q, k, v, o). Training uses learning rate $2 \times 10^{-4}$ with 3\% warmup, and bfloat16 precision. We train all models for one epoch.

\textbf{Instruction Tuning Data.} We compare TempoBench against three established open-source training datasets, each representing a different reasoning domain: \textbf{SlimOrca} (SO)~\cite{mukherjee2023orcaprogressivelearningcomplex}, a general-purpose instruction-following dataset distilled from GPT-4 explanation traces; \textbf{OpenMathInstruct v1} (OMI-1)~\cite{toshniwal2024openmathinstruct}, a large-scale mathematical reasoning dataset; and \textbf{OpenCoder} (OC)~\cite{huang2025opencoder}, a code-focused dataset for training code LLMs. Our data is much longer and denser than other reasoning datasets. We thus also train on a subsampled \textsc{TempoBench} dataset that matches the token counts with our comparison baselines (tb\_2k). For mixed-domain training, we combine the specialized datasets (50k + 50k = 100k samples total). We set a small number of \textsc{TempoBench} samples to supervise on CoT and JSON output, while the majority of samples mask the JSON output and supervise only the CoT.

\textbf{Evaluation Metrics.} For SIM, we report \textit{step accuracy}: the fraction of steps where the model correctly predicts the output and next state. For MIN, we report step accuracy and \textit{exact match} (EM; all steps in a sample correct). Each step corresponds to one transition in the execution trace (see Appendix~\ref{app:prompts} for the full prompt format). Responses that fail to parse into the required structured format are scored as zero (harsh denominator), ensuring that format compliance is not conflated with reasoning ability. We report MIN results at three granularities. MIN includes all causal steps, including ``no constraints'' (NC) steps where no input is causally necessary, so a model can achieve non-trivial MIN accuracy by defaulting to NC for every step. This is only an issue for finetuned models, since they may have learned this strategy. Identifying no constraints is hard and requires models to soundly perform counterfactual reasoning. MIN+ restricts evaluation to constrained steps only, where the gold answer requires identifying specific input literals; this is the primary metric for causal reasoning. MIN-Hard uses a separate NC-free evaluation set in which every step requires specific literals, eliminating the NC shortcut entirely. Finetuned models are evaluated with 4k max new tokens.

\textbf{Evaluation Benchmarks} We evaluate all finetuned models on a series of benchmarks that span reasoning domains such as mathematics, word problem reasoning, and logic. We provide descriptions of the benchmarks in Appendix~\ref{app:benchmarks}.

\section{Results and Discussion}
\label{sec:results}

\subsection{Main Results}
\begin{table*}[t]
  \centering
  \setlength{\tabcolsep}{2.5pt}
  \small
  \renewcommand{\arraystretch}{0.92}
  \resizebox{\textwidth}{!}{%
  \begin{tabular}{l|cccccccc|cccc}
    \toprule
    & & & & & & & & & \multicolumn{4}{c}{\textbf{\textsc{TempoBench}}} \\
    \textbf{Model} & ARC-C & WinoGr & GSM8K & MMLU & HellaS & MINERVA & IFEval & BBH-CoT & \textbf{SIM} & \textbf{MIN} & \textbf{MIN+} & \textbf{MIN-H} \\
    \midrule
    & \multicolumn{12}{c}{\textit{\textsc{TempoBench} transfers to standard reasoning benchmarks - other reasoning data does not transfer well}} \\
    \midrule
    \multicolumn{13}{l}{\textit{Llama-3.1-8B}} \\
    \cmidrule(l){1-13}
    baseline         & 52.0 & 74.3 & 51.0 & 55.4 & 60.7 & 19.0 & 9.1 & 64.7 & 0.0 & 0.0 & 0.0 & 0.0 \\
    SO   & \mgcell{55.8} & \lgcell{76.1} & \dgcell{57.3} & \lrcell{54.1} & \lgcell{62.0} & \mrcell{14.7} & \dgcell{23.5} & \mgcell{67.0} & \dgcell{27.0} & \mgcell{5.5} & \mgcell{9.3} & 0.0 \\
    OMI-1       & \lgcell{53.0} & \lrcell{72.5} & \dgcell{65.0} & \lgcell{55.5} & \lrcell{60.5} & \drcell{7.8} & \dgcell{14.8} & \lrcell{64.6} & \mgcell{6.0} & 0.4 & 0.0 & 0.0 \\
    OC       & \mgcell{54.1} & \lrcell{73.9} & \lrcell{49.7} & \lrcell{55.0} & \lgcell{61.1} & \lrcell{18.8} & \mgcell{12.9} & 64.7 & \dgcell{23.8} & 0.4 & \lgcell{2.1} & 0.0 \\
    OMI-1+SO       & \lgcell{53.8} & \lrcell{74.2} & \dgcell{64.7} & \lrcell{53.6} & \lgcell{61.0} & \drcell{11.1} & \dgcell{25.3} & \lgcell{65.1} & \dgcell{20.2} & \mgcell{6.9} & \lgcell{2.4} & 0.0 \\
    OC+SO       & \mgcell{56.1} & \lgcell{75.2} & \lgcell{52.8} & \lrcell{53.8} & \lgcell{61.5} & \lrcell{17.1} & \dgcell{23.1} & \mgcell{67.3} & \dgcell{20.4} & \dgcell{17.2} & \dgcell{13.3} & 0.0 \\
    \cmidrule(l){1-13}
    tb\_only & \lgcell{53.8} & \lrcell{73.8} & \lrcell{49.3} & \lrcell{53.9} & \mgcell{62.8} & \lrcell{18.4} & \lgcell{9.4} & \mrcell{62.3} & \dgcell{55.3} & \dgcell{33.5} & \mgcell{3.7} & 0.0 \\
    tb\_2k        & \mgcell{\textbf{56.2}} & \lrcell{73.2} & \lgcell{52.7} & \dgcell{\textbf{63.9}} & \dgcell{\textbf{80.4}} & \drcell{12.3} & \dgcell{15.6} & \lrcell{63.9} & \mgcell{2.9} & \lrcell{7.7} & \lgcell{0.8} & \dgcell{11.1} \\
    tb+SO         & \mgcell{55.5} & \lgcell{\textbf{75.2}} & \mgcell{56.0} & \lrcell{53.8} & \lgcell{61.9} & \lrcell{18.1} & \dgcell{\textbf{19.8}} & \lgcell{\textbf{65.9}} & \dgcell{64.7} & \dgcell{25.1} & \dgcell{\textbf{14.9}} & \dgcell{\textbf{11.9}} \\
    tb+OMI-1         & \mgcell{56.1} & \lrcell{72.9} & \dgcell{\textbf{63.8}} & \dgcell{\textbf{63.9}} & \dgcell{78.8} & \dgcell{\textbf{24.8}} & \dgcell{15.3} & \lgcell{65.2} & \dgcell{\textbf{72.8}} & \dgcell{35.2} & \mgcell{4.9} & 0.4 \\
    tb+OC         & \mgcell{54.4} & \lrcell{74.0} & \mgcell{55.3} & \dgcell{64.3} & \dgcell{79.9} & \mgcell{21.9} & \mgcell{12.4} & \lgcell{65.4} & \dgcell{35.5} & \dgcell{\textbf{41.1}} & \mgcell{6.3} & \lgcell{2.6} \\
    \midrule
    \multicolumn{13}{l}{\textit{Qwen2.5-7B}} \\
    \cmidrule(l){1-13}
    baseline         & 51.0 & 73.2 & 84.4 & 71.8 & 79.0 & 36.2 & 28.3 & 69.0 & 0.5 & 0.0 & 3.2 & 0.0 \\
    SO   & \mgcell{54.0} & \lgcell{73.4} & \lrcell{84.0} & 71.8 & \lrcell{78.9} & \lgcell{36.7} & \lgcell{29.9} & \lrcell{68.7} & \mgcell{7.7} & \mgcell{7.7} & 3.3 & 0.0 \\
    OMI-1       & \lrcell{50.6} & \lrcell{71.7} & \drcell{70.3} & \lrcell{71.5} & 79.0 & \drcell{15.6} & \mrcell{25.1} & \lrcell{68.0} & \mgcell{4.5} & \mgcell{3.3} & \mgcell{10.5} & 0.0 \\
    OC       & \mgcell{53.7} & \lrcell{72.8} & \lrcell{84.0} & \lgcell{71.9} & \lrcell{78.6} & \lrcell{34.5} & \lrcell{27.7} & \lrcell{67.8} & \mgcell{9.3} & 0.1 & \lrcell{0.7} & 0.0 \\
    OMI-1+SO       & \mgcell{55.1} & \lgcell{73.4} & \lrcell{83.9} & \lrcell{71.6} & \lrcell{78.9} & 36.2 & \lgcell{29.4} & \lgcell{69.1} & \mgcell{10.5} & \dgcell{16.8} & \mgcell{8.6} & 0.0 \\
    OC+SO       & \mgcell{55.2} & \lrcell{73.1} & \lrcell{83.4} & 71.8 & \lrcell{78.8} & \lgcell{37.4} & \lgcell{28.8} & \lgcell{69.1} & \mgcell{7.2} & \dgcell{13.6} & \lrcell{2.1} & 0.0 \\
    \cmidrule(l){1-13}
    tb\_only & \lgcell{52.8} & \lrcell{72.3} & \mrcell{81.1} & \lgcell{\textbf{71.9}} & \lgcell{\textbf{79.1}} & \dgcell{\textbf{42.3}} & \lrcell{27.9} & \mrcell{64.2} & \dgcell{\textbf{69.5}} & \dgcell{\textbf{25.8}} & \dgcell{\textbf{16.8}} & 0.0 \\
    tb\_2k        & \lgcell{51.2} & \mrcell{70.5} & \drcell{53.1} & \mrcell{68.3} & \drcell{58.5} & \drcell{22.7} & \lrcell{22.7} & \drcell{53.0} & \dgcell{5.0} & \dgcell{18.3} & \lgcell{6.9} & 0.0 \\
    tb+SO         & \mgcell{\textbf{54.3}} & 73.2 & \lrcell{84.2} & \lrcell{71.7} & \lrcell{78.8} & \mgcell{40.6} & \mgcell{\textbf{31.8}} & \lrcell{68.3} & \dgcell{14.3} & \mgcell{3.5} & \mgcell{7.1} & 0.0 \\
    tb+OMI-1         & \lgcell{51.7} & \mrcell{71.2} & \mrcell{82.4} & \lrcell{71.6} & 79.0 & \drcell{27.3} & \lgcell{29.9} & \mrcell{65.9} & \dgcell{75.0} & \dgcell{22.4} & \mgcell{7.0} & 0.1 \\
    tb+OC         & \mgcell{53.6} & \lrcell{72.5} & \lrcell{83.2} & \lgcell{\textbf{71.9}} & \lrcell{78.9} & \mgcell{39.9} & \lgcell{29.2} & \mrcell{66.7} & \dgcell{53.0} & \dgcell{20.0} & \mgcell{12.1} & 0.0 \\
    \midrule
    \multicolumn{13}{l}{\textit{Llama-3.2-3B}} \\
    \cmidrule(l){1-13}
    baseline         & 46.4 & 69.8 & 30.1 & 55.0 & 74.1 & 6.3 & 7.8 & 47.6 & 0.0 & 0.0 & 0.0 & 0.0 \\
    SO   & \lgcell{47.7} & \mgcell{71.9} & \mgcell{33.4} & \lrcell{53.4} & \lgcell{74.4} & \lgcell{6.6} & \dgcell{14.2} & \lgcell{48.1} & \mgcell{3.6} & \graycell{78.2} & 0.0 & 0.0 \\
    OMI-1       & \lrcell{44.8} & \lgcell{69.9} & \dgcell{46.2} & \lrcell{53.9} & \lrcell{72.3} & \mrcell{3.5} & \dgcell{12.8} & \lrcell{47.5} & \lgcell{0.6} & \lgcell{1.1} & 0.0 & 0.0 \\
    OC       & \lgcell{46.5} & \lgcell{70.1} & \lrcell{28.4} & \lrcell{54.8} & \lgcell{74.2} & \lrcell{4.8} & \lgcell{9.1} & \lrcell{46.8} & 0.1 & \dgcell{18.4} & 0.0 & 0.0 \\
    OMI-1+SO       & \lgcell{48.2} & \lgcell{71.2} & \dgcell{47.8} & \lrcell{54.1} & \lrcell{73.6} & \lgcell{7.4} & \dgcell{15.5} & \drcell{28.5} & \lgcell{1.9} & \graycell{78.2} & 0.0 & 0.0 \\
    OC+SO       & \lgcell{46.8} & \lgcell{71.1} & \lgcell{31.9} & \lrcell{54.1} & \lgcell{74.2} & \lgcell{6.6} & \dgcell{15.5} & \drcell{33.7} & \lgcell{1.5} & \graycell{75.0} & \lgcell{0.7} & 0.0 \\
    \cmidrule(l){1-13}
    tb\_only & \lgcell{46.8} & \lrcell{69.4} & \mrcell{27.6} & \lrcell{54.7} & \lgcell{\textbf{75.4}} & \lgcell{7.3} & \dgcell{13.3} & \mrcell{43.0} & \dgcell{23.7} & \dgcell{\textbf{21.5}} & \dgcell{\textbf{13.9}} & \lgcell{\textbf{1.5}} \\
    tb\_2k        & \lrcell{44.5} & \lgcell{70.3} & \drcell{12.7} & \mrcell{50.3} & \lrcell{73.4} & \mrcell{2.8} & \dgcell{\textbf{15.5}} & \drcell{15.3} & \lgcell{1.6} & \drcell{14.7} & 0.0 & 0.0 \\
    tb+SO         & \mgcell{\textbf{48.6}} & \lgcell{\textbf{71.4}} & \mgcell{33.1} & \lrcell{54.0} & \lgcell{74.6} & \lgcell{7.2} & \dgcell{12.9} & \drcell{24.0} & \dgcell{48.1} & \dgcell{16.9} & \mgcell{3.1} & 0.0 \\
    tb+OMI-1         & \lrcell{45.5} & \lgcell{70.0} & \dgcell{\textbf{47.3}} & \lrcell{54.5} & \lrcell{73.8} & \lgcell{\textbf{7.5}} & \mgcell{11.1} & \lrcell{47.0} & \dgcell{\textbf{53.9}} & \dgcell{24.2} & \lgcell{3.0} & 0.0 \\
    tb+OC         & \lrcell{46.1} & \lgcell{70.4} & \mrcell{27.1} & \lrcell{54.5} & \lgcell{74.5} & 6.3 & \mgcell{11.7} & \lrcell{47.5} & \mgcell{3.8} & \dgcell{12.2} & \lgcell{2.2} & 0.3 \\
    \bottomrule
  \end{tabular}}
  \renewcommand{\arraystretch}{1.0}
  \caption{Main evaluation: \textsc{TempoBench training generalizes; other data does not teach causal reasoning} (sample-level averages; unparsed = 0). Left: standard benchmarks. Right: \textsc{TempoBench} eval. Green/red cells denote improvements/declines relative to baseline. Bold: best per column within sub-group. Gray cells denote cases where the model only predicts NC.}
  \label{tab:main_results}
\end{table*}

Table~\ref{tab:llm_comp} presents frontier and fine-tuned model results on \textsc{TempoBench}. On SIM, all closed source models achieve step accuracy ranging from 23.5\% to 96.0\%. However, on MIN two of four frontier models score below 1\%, and even on constrained steps (MIN+), the best frontier model (Haiku 4.5) reaches only 24.4\% (demonstrating that NC steps are not trivial in practice). The high overspecification rates show that frontier models list unnecessary literals on 60--96\% of MIN steps; even with the algorithm, the models prefer to overspecify. No model exceeds 29\% on constrained causal steps, demonstrating a general limitation, not a prompting limitation. We show that much smaller task fine-tuned open-source models can match or exceed frontier models on the \textsc{TempoBench} evals. LLaMA 3.1 8B with TB+SO training achieves 21.7\% on constrained causal steps (MIN+), exceeding Sonnet~4.6's 13.7\% and approaching Haiku~4.5's 24.4\%. Qwen 2.5 7B with TB+OMI-1 reaches 22.4\% on MIN, surpassing three of the frontier models. This demonstrates that temporal causal reasoning is a learnable capability that can be acquired from synthetic formally verified data, and is harder to acquire from standard reasoning training (math, code, or instruction data). We find that closed source reasoning models run with up to 20k token usage are able to improve performance. Despite overspecifying more on the hard task, DeepSeek V3.2-reasoning improves significantly over its non reasoning counterpart (we conduct a detailed analysis of model behavioral differences in Appendix~\ref{app:complexity-sweep}).

Table~\ref{tab:main_results} shows downstream benchmark results across all three model families. \textsc{TempoBench} training provides a more balanced signal than any single-domain alternative. Qwen 7B, LLaMA 3B degrade more under the smaller tb\_2k, similar to training on OMI-1, but LLaMA 8B improves greatly across all evals. For LLaMA 3.1 8B, mixing \textsc{TempoBench} with other data produces the strongest overall models of any training condition: tb+OMI-1 averages $+6.8$ points across all standard benchmarks (HellaSwag $+18.1$, MMLU $+8.5$, GSM8K $+12.8$, MINERVA $+5.8$), and tb+OC averages $+5.2$ (HellaSwag $+19.2$, MMLU $+8.9$), exceeding SO ($+3.0$ average), which is a general instruction-following dataset. In contrast, OMI-1-only training gains $+14$ on GSM8K but loses $-11.2$ on MINERVA, demonstrating severe overfitting. The very strong improvements in the HellaS and IFEval benchmarks can be attributed to the enhanced long horizon CoT adherence that this data may train. The \textsc{TempoBench} reasoning format appears to be more generalizable and transfer more broadly, improving performance across the board. For Qwen 2.5 7B and LLaMA 3B, gains are smaller, but we observe patterns comparable to the other training datasets. This pattern is amplified with the tb\_2k dataset, where Qwen 7B and LLaMA 3B show higher average degradation than LLaMA 8B. These two models, being more heavily optimized and of a smaller scale, appear more fragile to over or undertraining. Ultimately, across all three model families, and in particular with the 8B parameter model, training with \textsc{TempoBench} yields the most balanced gains, clearly showing that the data generalizes well to other tasks, while other tasks do not necessarily transfer well to the \textsc{TempoBench} data. Overall, we find that mixing \textsc{TempoBench} with other datasets yields the best results.

\subsection{Brute Forcing Reasoning Patterns}
\begin{table*}[h]
  \centering
  \small
  \setlength{\tabcolsep}{2.5pt}
  \resizebox{\textwidth}{!}{%
  \begin{tabular}{l|cccc|cccc|cccc}
    \toprule
    & \multicolumn{4}{c|}{\textbf{MIN}} & \multicolumn{4}{c|}{\textbf{MIN+}} & \multicolumn{4}{c}{\textbf{MIN-Hard}} \\
    \textbf{Model} & Correct$\uparrow$ & Ovsp$\downarrow$ & C.Copy$\downarrow$ & Unsp$\downarrow$ & Correct$\uparrow$ & Ovsp$\downarrow$ & C.Copy$\downarrow$ & Unsp$\downarrow$ & Correct$\uparrow$ & Ovsp$\downarrow$ & C.Copy$\downarrow$ & Unsp$\downarrow$ \\
    \midrule
    \multicolumn{13}{l}{\textit{Frontier models}} \\
    \cmidrule(l){1-13}
    Sonnet 4.6         & 36.8 & 60.1 & 49.1 & 3.2 & 14.0 & 69.7 & 13.3 & 16.3 &  5.4 & 80.9 & 10.4 & 13.8 \\
    Haiku 4.5          &  0.9 & 98.9 & 93.0 & 0.2 &  0.6 & 97.9 & 53.3 &  1.5 &  1.2 & 98.3 & 93.1 &  0.5 \\
    DeepSeek V3.2 (normal)&  0.1 & 99.5 & 93.3 & 0.0 & 0.4 & 99.4 & 53.3 & 0.0 & 0.0 & 98.8 & 93.3 & 0.0 \\
    DeepSeek V3        & 20.4 & 76.8 & 60.6 & 2.8 &  2.7 & 83.6 &  3.9 & 13.7 &  2.3 & 84.2 &  1.2 & 13.5 \\
    \cmidrule(l){1-13}
    \multicolumn{13}{l}{\textit{Reasoning models}} \\
    \cmidrule(l){1-13}
    DeepSeek R1            & 63.1 & 27.6 & 23.3 & 5.1 & 10.3 & 48.2 & 16.1 & 38.8 & 59.0 & 32.1 & 29.4 & 5.2 \\
    DeepSeek V3.2 (reason) & 55.2 & 14.9 & 12.0 & 5.0 &  9.7 & 30.9 &  9.0 & 37.8 & 53.3 & 16.2 & 14.2 & 4.8 \\
    \midrule
    \multicolumn{13}{l}{\textit{Best TB-trained per family}} \\
    \cmidrule(l){1-13}
    8B tb+SO       & 26.7 & 66.8 & 63.2 &  6.5 & 14.9 & 49.2 & 29.4 & 35.9 & 12.0 & 38.4 &  2.7 & 49.6 \\
    7B tb+OMI-1    & 27.3 & 67.6 & 58.8 &  5.1 &  7.0 & 66.3 & 20.5 & 26.7 &  0.1 & 27.8 &  0.1 & 72.1 \\
    3B tb+OC       & 14.8 & 81.4 & 69.9 &  3.8 &  2.2 & 76.3 & 11.8 & 21.5 &  2.1 & 35.0 &  4.2 & 62.9 \\
    \bottomrule
  \end{tabular}}
  \caption{\textbf{Error type breakdown across MIN difficulty.} (step-level:  aggregates steps across all samples; unparsed scored as 0). \textit{Correct}: FP${=}$0, FN${=}$0. \textit{Ovsp}: FP${>}$0 (unnecessary literals). \textit{C.Copy}: FP${>}$0, FN${=}$0 (pure condition retrieval). \textit{Unsp}: FP${=}$0, FN${>}$0 (missed necessary literals).}
  \label{tab:retrieval}
\end{table*}

Table~\ref{tab:retrieval} showcases model retrieval patterns on MIN task causal effect identification tasks. Additional metrics decompose errors on the causal reasoning task: \textit{Condition Copy} (FP${>}$0, FN${=}$0) indicates the model listed unnecessary literals but missed none, consistent with copying the full transition condition verbatim. \textit{Underspec} (FP${=}$0, FN${>}$0) indicates the model missed necessary literals without adding extras, consistent with attempted reasoning that fell short. Frontier models show very high condition copy rates across all of the eval sets, indicating that they are biased towards retrieval. On the easier evaluation set (which includes NC steps), the frontier models perform better, confirming that the retrieval strategy succeeds on trivial steps but produces incoherent outputs on genuinely constrained ones. We also examine whether extended inference-time reasoning improves causal attribution. Table~\ref{tab:retrieval} includes DeepSeek R1 and V3.2 in reasoning mode, which consume around 20k tokens per sample. Extended reasoning generally reduces condition copy within a family but tends to raise underspecification. Additional inference compute does not address the fundamental causal attribution gap. Reasoning models have the lowest copy rate on the MIN task, but copy more and perform worse on the MIN+ and MIN-Hard tasks, indicating that the models are reasoning and trying to exploit structural difficulties with these eval sets (Appendix~\ref{app:complexity-sweep}). The TB-trained models show a different pattern: low condition copy (2.7\%) and high underspecification (49.6\%) on MIN-Hards, indicating they are attempting to perform counterfactual reasoning but miss some necessary literals. We provide a statistical evaluation of how dataset features like trace length and transition count influence model performance in Appendix~\ref{app:complexity-sweep}. Our results indicate that low performance on the MIN task is not bottlenecked by task complexity.

\subsection{Effect of Causal CoT}
\begin{table*}[h]
  \centering
  \small
  \setlength{\tabcolsep}{2.5pt}
  \resizebox{\textwidth}{!}{%
  \begin{tabular}{l|cccccccc|cccc}
    \toprule
    & & & & & & & & & \multicolumn{4}{c}{\textbf{\textsc{TempoBench}}} \\
    \textbf{Model} & ARC-C & WinoGr & GSM8K & MMLU & HellaS & MINERVA & IFEval & BBH-CoT & \textbf{SIM} & \textbf{MIN} & \textbf{MIN+} & \textbf{MIN-H} \\
    \midrule
    baseline         & 52.0 & 74.3 & 51.0 & 55.4 & 60.7 & 19.0 & 9.1 & 64.7 & 0.3 & 8.7 & 0.0 & 0.0 \\
    OC+SO        & 56.1 & 75.2 & 52.8 & 53.8 & 61.5 & 17.1 & 23.1 & 67.3 & 3.5 & 75.1 & 15.4 & 0.0 \\
    \midrule
    tb (50k, with CoT)    & 53.8 & 73.8 & 49.3 & 53.9 & 62.8 & 18.4 & 9.4 & 62.3 & 19.9 & 25.6 & 15.6 & 0.0 \\
    \hspace{2mm}\textit{$\Delta$ baseline} & \textit{+1.8} & \textit{$-$0.5} & \textit{$-$1.7} & \textit{$-$1.5} & \textit{+2.1} & \textit{$-$0.6} & \textit{+0.3} & \textit{$-$2.4} & \textit{+19.6} & \textit{+16.9} & \textit{+15.6} & \textit{0.0} \\
    tb (50k, no CoT)      & 52.2 & 71.5 & 28.9 & 58.2 & 78.6 & 8.0 & 8.3 & 42.0 & 0.0 & 11.9 & 0.0 & 0.0 \\
    \hspace{2mm}\textit{$\Delta$ baseline} & \textit{+0.2} & \textit{$-$2.8} & \textit{$-$22.1} & \textit{+2.8} & \textit{+17.9} & \textit{$-$11.0} & \textit{$-$0.8} & \textit{$-$22.7} & \textit{$-$0.3} & \textit{+3.2} & \textit{0.0} & \textit{0.0} \\
    \bottomrule
  \end{tabular}}
  \caption{\textbf{CoT Inclusion Ablation} on Llama-3.1-8B trained with and without CoT supervision.}
  \label{tab:nocot}
\end{table*}

We evaluate the importance of the inclusion of the CoT reasoning traces explaining the temporal causality in the systems as shown in Figure~\ref{fig:pipeline} and in Appendix~\ref{app:prompts}. Instead, we only train the model using the automaton description as input, and the final answers as supervision targets. We train the Llama 8B model for this ablation. As shown in Table~\ref{tab:nocot}, training without the CoT supervision seriously harms model performance on almost all benchmarks compared to baseline and tempobench trained results. We see a particular drop in benchmarks that require multi-turn reasoning, supporting the idea that it is the actual temporal causal reasoning that generalizes well to other tasks. We see a slight uplift in the MIN task over the baseline, which is likely due to the model getting some weak learning signal from the no CoT data.

\subsection{Testing for memorization}
\begin{table*}[h]
  \centering
  \small
  \setlength{\tabcolsep}{2.5pt}
  \resizebox{\textwidth}{!}{%
  \begin{tabular}{l|ccc|cccc|cccc|cccc}
    \toprule
    & \multicolumn{3}{c|}{\textbf{SIM}}
    & \multicolumn{4}{c|}{\textbf{MIN}}
    & \multicolumn{4}{c|}{\textbf{MIN+}}
    & \multicolumn{4}{c}{\textbf{MIN-Hard}} \\
    \textbf{Visible}
    & Parse & Step & EM
    & Parse & Step & EM & C.Copy$\downarrow$
    & Parse & Step & EM & C.Copy$\downarrow$
    & Parse & Step & EM & C.Copy$\downarrow$ \\
    \midrule
    100\%  & 95.9 & 59.8 & 26.0 & 95.9 & 36.3 & 15.5 & 48.1 & 95.9 & 15.6 & 16.6 & 19.8 & 94.0 & 0.0 & 0.0 & 4.8 \\
    50\%   & 94.5 & 21.7 &  0.9 & 94.5 & 38.3 & 15.3 & 45.2 & 94.5 & 12.6 & 16.9 & 20.4 & 93.1 & 0.0 & 0.0 & 6.9 \\
    25\%   & 94.5 &  8.8 &  0.0 & 94.5 & 30.3 &  9.2 & 54.5 & 94.5 & 12.2 & 14.5 & 22.8 & 93.5 & 0.1 & 0.0 & 7.9 \\
    \bottomrule
  \end{tabular}}
  \caption{\textbf{Transition visibility ablation} tempobench\_only, LLaMA-3.1-8B.}
  \label{tab:memorization_test}
\end{table*}

We perform an additional ablation to test whether models are memorizing the machines from the syntcomp benchmarks. We test on SIM, MIN, and MIN-Hard, with $100\%,50\%,25\%$ of the input visible to evaluate how much the models are memorizing or copying from the input trace, versus actually reasoning over the input. As shown in Table~\ref{tab:memorization_test}, SIM degrades sharply and proportionally to the degree of transition obfuscation (59.8\% $\to$ 8.8\%), confirming the model reads and correctly uses the machine for trace simulation. MIN is less affected (36.3\% $\to$ 30.3\%); however, condition copy increases (48.1\% $\to$ 54.5\%). MIN+ shows a similar pattern to MIN. For all variations of MIN, we see a tendency of the model to increasingly copy from the transition conditions as visibility reduces. Qualitative analysis of the models' outputs shown in Appendix~\ref{app:truncated-output-samples} illustrates the argument that the models are guessing and going in circles trying to reason about the input. This behavior indicates that the model is referencing the input system when producing its CoT, rather than memorizing patterns. 

\section{Conclusion}
We introduced \textsc{TempoBench}, a formally verified benchmark for temporal reasoning over execution traces, and demonstrated that automata offer a way to generate scalable, high-quality, verified reasoning traces. We also performed a systematic evaluation of LLMs showing fundamental challenges with reasoning questions that require models to reason and return \textit{less} information than is contained in the input, as would be the case in root cause analysis. Fine-tuning on \textsc{TempoBench} improves this capability while preserving, and often improving, performance on standard benchmarks, suggesting that reasoning over system behavior transfers beyond the benchmark to broader domains. Our results indicate that this form of reasoning is a learnable capability enabled by the right training signal, rather than an emergent property of scale alone.

\bibliography{colm2026_conference}

@inproceedings{coenen2022temporal,
  title={Temporal causality in reactive systems},
  author={Coenen, Norine and Finkbeiner, Bernd and Frenkel, Hadar and Hahn, Christopher and Metzger, Niklas and Siber, Julian},
  booktitle={International symposium on automated technology for verification and analysis},
  pages={208--224},
  year={2022},
  organization={Springer}
}

@article{ikram2022root,
  title={Root cause analysis of failures in microservices through causal discovery},
  author={Ikram, Azam and Chakraborty, Sarthak and Mitra, Subrata and Saini, Shiv and Bagchi, Saurabh and Kocaoglu, Murat},
  journal={Advances in Neural Information Processing Systems},
  volume={35},
  pages={31158--31170},
  year={2022}
}

@inproceedings{finkbeiner2024synthesis,
  title={Synthesis of temporal causality},
  author={Finkbeiner, Bernd and Frenkel, Hadar and Metzger, Niklas and Siber, Julian},
  booktitle={International Conference on Computer Aided Verification},
  pages={87--111},
  year={2024},
  organization={Springer}
}

@inproceedings{pnueli1977temporal,
  title={The temporal logic of programs},
  author={Pnueli, Amir},
  booktitle={18th annual symposium on foundations of computer science (sfcs 1977)},
  pages={46--57},
  year={1977},
  organization={ieee}
}

@inproceedings{pnueli1989synthesis,
  title={On the synthesis of a reactive module},
  author={Pnueli, Amir and Rosner, Roni},
  booktitle={Proceedings of the 16th ACM SIGPLAN-SIGACT symposium on Principles of programming languages},
  pages={179--190},
  year={1989}
}

@article{church1963application,
  title={Application of recursive arithmetic to the problem of circuit synthesis},
  author={Church, Alonzo},
  journal={Journal of Symbolic Logic},
  volume={28},
  number={4},
  year={1963}
}

@inproceedings{finkbeiner2019temporal,
  title={Temporal stream logic: Synthesis beyond the bools},
  author={Finkbeiner, Bernd and Klein, Felix and Piskac, Ruzica and Santolucito, Mark},
  booktitle={International Conference on Computer Aided Verification},
  pages={609--629},
  year={2019},
  organization={Springer}
}

@misc{chollet2025arcprize2024technical,
      title={ARC Prize 2024: Technical Report}, 
      author={Francois Chollet and Mike Knoop and Gregory Kamradt and Bryan Landers},
      year={2025},
      eprint={2412.04604},
      archivePrefix={arXiv},
      primaryClass={cs.AI},
      url={https://arxiv.org/abs/2412.04604}, 
}

@misc{chollet2025arcagi2newchallengefrontier,
      title={ARC-AGI-2: A New Challenge for Frontier AI Reasoning Systems}, 
      author={Francois Chollet and Mike Knoop and Gregory Kamradt and Bryan Landers and Henry Pinkard},
      year={2025},
      eprint={2505.11831},
      archivePrefix={arXiv},
      primaryClass={cs.AI},
      url={https://arxiv.org/abs/2505.11831}, 
}

@article{leucker2009brief,
  title={A brief account of runtime verification},
  author={Leucker, Martin and Schallhart, Christian},
  journal={The journal of logic and algebraic programming},
  volume={78},
  number={5},
  pages={293--303},
  year={2009},
  publisher={Elsevier}
}

@article{ding2025understanding,
  title={Understanding world or predicting future? a comprehensive survey of world models},
  author={Ding, Jingtao and Zhang, Yunke and Shang, Yu and Zhang, Yuheng and Zong, Zefang and Feng, Jie and Yuan, Yuan and Su, Hongyuan and Li, Nian and Sukiennik, Nicholas and others},
  journal={ACM Computing Surveys},
  volume={58},
  number={3},
  pages={1--38},
  year={2025},
  publisher={ACM New York, NY}
}

@misc{syntcomp_benchmarks,
  author       = {{SYNTCOMP}},
  title        = {{SYNTCOMP/benchmarks: Repository of benchmarks for the Reactive Synthesis Competition (SYNTCOMP)}},
  howpublished = {\url{https://github.com/SYNTCOMP/benchmarks}},
  note         = {Accessed: 2025-10-05},
  year         = {2025}
}

@inproceedings{DBLP:journals/corr/Jacobs016,
  author       = {Swen Jacobs and
                  Felix Klein and
                  Sebastian Schirmer},
  editor       = {Ruzica Piskac and
                  Rayna Dimitrova},
  title        = {A High-Level {LTL} Synthesis Format: {TLSF} v1.1},
  booktitle    = {Proceedings Fifth Workshop on Synthesis, SYNT@CAV 2016, Toronto, Canada,
                  July 17-18, 2016},
  series       = {{EPTCS}},
  pages        = {112--132},
  year         = {2016},
  url          = {https://doi.org/10.4204/EPTCS.229.10},
  doi          = {10.4204/EPTCS.229.10},
  bibsource    = {dblp computer science bibliography, https://dblp.org}
}

@article{renkin2022dissecting,
  title={Dissecting ltlsynt},
  author={Renkin, Florian and Schlehuber-Caissier, Philipp and Duret-Lutz, Alexandre and Pommellet, Adrien},
  journal={Formal Methods in System Design},
  volume={61},
  number={2},
  pages={248--289},
  year={2022},
  publisher={Springer}
}

@inproceedings{10.1007/978-3-032-05435-7_3,
author = {Di Stefano, Luca},
title = {Execution and Monitoring of HOA Automata with HOAX},
year = {2025},
isbn = {978-3-032-05434-0},
publisher = {Springer-Verlag},
address = {Berlin, Heidelberg},
url = {https://doi.org/10.1007/978-3-032-05435-7_3},
doi = {10.1007/978-3-032-05435-7_3},
booktitle = {Runtime Verification: 25th International Conference, RV 2025, Graz, Austria, September 15–19, 2025, Proceedings},
pages = {44–53},
numpages = {10},
location = {Graz, Austria}
}

@inproceedings{
fatemi2025test,
title={Test of Time: A Benchmark for Evaluating {LLM}s on Temporal Reasoning},
author={Bahare Fatemi and Mehran Kazemi and Anton Tsitsulin and Karishma Malkan and Jinyeong Yim and John Palowitch and Sungyong Seo and Jonathan Halcrow and Bryan Perozzi},
booktitle={The Thirteenth International Conference on Learning Representations},
year={2025},
url={https://openreview.net/forum?id=44CoQe6VCq}
}

@inproceedings{
jimenez2024swebench,
title={{SWE}-bench: Can Language Models Resolve Real-world Github Issues?},
author={Carlos E Jimenez and John Yang and Alexander Wettig and Shunyu Yao and Kexin Pei and Ofir Press and Karthik R Narasimhan},
booktitle={The Twelfth International Conference on Learning Representations},
year={2024},
url={https://openreview.net/forum?id=VTF8yNQM66}
}

@misc{cobbe2021trainingverifierssolvemath,
      title={Training Verifiers to Solve Math Word Problems}, 
      author={Karl Cobbe and Vineet Kosaraju and Mohammad Bavarian and Mark Chen and Heewoo Jun and Lukasz Kaiser and Matthias Plappert and Jerry Tworek and Jacob Hilton and Reiichiro Nakano and Christopher Hesse and John Schulman},
      year={2021},
      eprint={2110.14168},
      archivePrefix={arXiv},
      primaryClass={cs.LG},
      url={https://arxiv.org/abs/2110.14168}, 
}

@misc{chen2021evaluatinglargelanguagemodels,
      title={Evaluating Large Language Models Trained on Code}, 
      author={Mark Chen and Jerry Tworek and Heewoo Jun and Qiming Yuan and Henrique Ponde de Oliveira Pinto and Jared Kaplan and Harri Edwards and Yuri Burda and Nicholas Joseph and Greg Brockman and Alex Ray and Raul Puri and Gretchen Krueger and Michael Petrov and Heidy Khlaaf and Girish Sastry and Pamela Mishkin and Brooke Chan and Scott Gray and Nick Ryder and Mikhail Pavlov and Alethea Power and Lukasz Kaiser and Mohammad Bavarian and Clemens Winter and Philippe Tillet and Felipe Petroski Such and Dave Cummings and Matthias Plappert and Fotios Chantzis and Elizabeth Barnes and Ariel Herbert-Voss and William Hebgen Guss and Alex Nichol and Alex Paino and Nikolas Tezak and Jie Tang and Igor Babuschkin and Suchir Balaji and Shantanu Jain and William Saunders and Christopher Hesse and Andrew N. Carr and Jan Leike and Josh Achiam and Vedant Misra and Evan Morikawa and Alec Radford and Matthew Knight and Miles Brundage and Mira Murati and Katie Mayer and Peter Welinder and Bob McGrew and Dario Amodei and Sam McCandlish and Ilya Sutskever and Wojciech Zaremba},
      year={2021},
      eprint={2107.03374},
      archivePrefix={arXiv},
      primaryClass={cs.LG},
      url={https://arxiv.org/abs/2107.03374}, 
}

@article{
srivastava2023beyond,
title={Beyond the Imitation Game: Quantifying and extrapolating the capabilities of language models},
author={Aarohi Srivastava and Abhinav Rastogi and Abhishek Rao and Abu Awal Md Shoeb and Abubakar Abid and Adam Fisch and Adam R. Brown and Adam Santoro and Aditya Gupta and Adri{\`a} Garriga-Alonso and Agnieszka Kluska and Aitor Lewkowycz and Akshat Agarwal and Alethea Power and Alex Ray and Alex Warstadt and Alexander W. Kocurek and Ali Safaya and Ali Tazarv and Alice Xiang and Alicia Parrish and Allen Nie and Aman Hussain and Amanda Askell and Amanda Dsouza and Ambrose Slone and Ameet Rahane and Anantharaman S. Iyer and Anders Johan Andreassen and Andrea Madotto and Andrea Santilli and Andreas Stuhlm{\"u}ller and Andrew M. Dai and Andrew La and Andrew Kyle Lampinen and Andy Zou and Angela Jiang and Angelica Chen and Anh Vuong and Animesh Gupta and Anna Gottardi and Antonio Norelli and Anu Venkatesh and Arash Gholamidavoodi and Arfa Tabassum and Arul Menezes and Arun Kirubarajan and Asher Mullokandov and Ashish Sabharwal and Austin Herrick and Avia Efrat and Aykut Erdem and Ayla Karaka{\c{s}} and B. Ryan Roberts and Bao Sheng Loe and Barret Zoph and Bart{\l}omiej Bojanowski and Batuhan {\"O}zyurt and Behnam Hedayatnia and Behnam Neyshabur and Benjamin Inden and Benno Stein and Berk Ekmekci and Bill Yuchen Lin and Blake Howald and Bryan Orinion and Cameron Diao and Cameron Dour and Catherine Stinson and Cedrick Argueta and Cesar Ferri and Chandan Singh and Charles Rathkopf and Chenlin Meng and Chitta Baral and Chiyu Wu and Chris Callison-Burch and Christopher Waites and Christian Voigt and Christopher D Manning and Christopher Potts and Cindy Ramirez and Clara E. Rivera and Clemencia Siro and Colin Raffel and Courtney Ashcraft and Cristina Garbacea and Damien Sileo and Dan Garrette and Dan Hendrycks and Dan Kilman and Dan Roth and C. Daniel Freeman and Daniel Khashabi and Daniel Levy and Daniel Mosegu{\'\i} Gonz{\'a}lez and Danielle Perszyk and Danny Hernandez and Danqi Chen and Daphne Ippolito and Dar Gilboa and David Dohan and David Drakard and David Jurgens and Debajyoti Datta and Deep Ganguli and Denis Emelin and Denis Kleyko and Deniz Yuret and Derek Chen and Derek Tam and Dieuwke Hupkes and Diganta Misra and Dilyar Buzan and Dimitri Coelho Mollo and Diyi Yang and Dong-Ho Lee and Dylan Schrader and Ekaterina Shutova and Ekin Dogus Cubuk and Elad Segal and Eleanor Hagerman and Elizabeth Barnes and Elizabeth Donoway and Ellie Pavlick and Emanuele Rodol{\`a} and Emma Lam and Eric Chu and Eric Tang and Erkut Erdem and Ernie Chang and Ethan A Chi and Ethan Dyer and Ethan Jerzak and Ethan Kim and Eunice Engefu Manyasi and Evgenii Zheltonozhskii and Fanyue Xia and Fatemeh Siar and Fernando Mart{\'\i}nez-Plumed and Francesca Happ{\'e} and Francois Chollet and Frieda Rong and Gaurav Mishra and Genta Indra Winata and Gerard de Melo and Germ{\`a}n Kruszewski and Giambattista Parascandolo and Giorgio Mariani and Gloria Xinyue Wang and Gonzalo Jaimovitch-Lopez and Gregor Betz and Guy Gur-Ari and Hana Galijasevic and Hannah Kim and Hannah Rashkin and Hannaneh Hajishirzi and Harsh Mehta and Hayden Bogar and Henry Francis Anthony Shevlin and Hinrich Schuetze and Hiromu Yakura and Hongming Zhang and Hugh Mee Wong and Ian Ng and Isaac Noble and Jaap Jumelet and Jack Geissinger and Jackson Kernion and Jacob Hilton and Jaehoon Lee and Jaime Fern{\'a}ndez Fisac and James B Simon and James Koppel and James Zheng and James Zou and Jan Kocon and Jana Thompson and Janelle Wingfield and Jared Kaplan and Jarema Radom and Jascha Sohl-Dickstein and Jason Phang and Jason Wei and Jason Yosinski and Jekaterina Novikova and Jelle Bosscher and Jennifer Marsh and Jeremy Kim and Jeroen Taal and Jesse Engel and Jesujoba Alabi and Jiacheng Xu and Jiaming Song and Jillian Tang and Joan Waweru and John Burden and John Miller and John U. Balis and Jonathan Batchelder and Jonathan Berant and J{\"o}rg Frohberg and Jos Rozen and Jose Hernandez-Orallo and Joseph Boudeman and Joseph Guerr and Joseph Jones and Joshua B. Tenenbaum and Joshua S. Rule and Joyce Chua and Kamil Kanclerz and Karen Livescu and Karl Krauth and Karthik Gopalakrishnan and Katerina Ignatyeva and Katja Markert and Kaustubh Dhole and Kevin Gimpel and Kevin Omondi and Kory Wallace Mathewson and Kristen Chiafullo and Ksenia Shkaruta and Kumar Shridhar and Kyle McDonell and Kyle Richardson and Laria Reynolds and Leo Gao and Li Zhang and Liam Dugan and Lianhui Qin and Lidia Contreras-Ochando and Louis-Philippe Morency and Luca Moschella and Lucas Lam and Lucy Noble and Ludwig Schmidt and Luheng He and Luis Oliveros-Col{\'o}n and Luke Metz and L{\"u}tfi Kerem Senel and Maarten Bosma and Maarten Sap and Maartje Ter Hoeve and Maheen Farooqi and Manaal Faruqui and Mantas Mazeika and Marco Baturan and Marco Marelli and Marco Maru and Maria Jose Ramirez-Quintana and Marie Tolkiehn and Mario Giulianelli and Martha Lewis and Martin Potthast and Matthew L Leavitt and Matthias Hagen and M{\'a}ty{\'a}s Schubert and Medina Orduna Baitemirova and Melody Arnaud and Melvin McElrath and Michael Andrew Yee and Michael Cohen and Michael Gu and Michael Ivanitskiy and Michael Starritt and Michael Strube and Micha{\l} Sw{\k{e}}drowski and Michele Bevilacqua and Michihiro Yasunaga and Mihir Kale and Mike Cain and Mimee Xu and Mirac Suzgun and Mitch Walker and Mo Tiwari and Mohit Bansal and Moin Aminnaseri and Mor Geva and Mozhdeh Gheini and Mukund Varma T and Nanyun Peng and Nathan Andrew Chi and Nayeon Lee and Neta Gur-Ari Krakover and Nicholas Cameron and Nicholas Roberts and Nick Doiron and Nicole Martinez and Nikita Nangia and Niklas Deckers and Niklas Muennighoff and Nitish Shirish Keskar and Niveditha S. Iyer and Noah Constant and Noah Fiedel and Nuan Wen and Oliver Zhang and Omar Agha and Omar Elbaghdadi and Omer Levy and Owain Evans and Pablo Antonio Moreno Casares and Parth Doshi and Pascale Fung and Paul Pu Liang and Paul Vicol and Pegah Alipoormolabashi and Peiyuan Liao and Percy Liang and Peter W Chang and Peter Eckersley and Phu Mon Htut and Pinyu Hwang and Piotr Mi{\l}kowski and Piyush Patil and Pouya Pezeshkpour and Priti Oli and Qiaozhu Mei and Qing Lyu and Qinlang Chen and Rabin Banjade and Rachel Etta Rudolph and Raefer Gabriel and Rahel Habacker and Ramon Risco and Rapha{\"e}l Milli{\`e}re and Rhythm Garg and Richard Barnes and Rif A. Saurous and Riku Arakawa and Robbe Raymaekers and Robert Frank and Rohan Sikand and Roman Novak and Roman Sitelew and Ronan Le Bras and Rosanne Liu and Rowan Jacobs and Rui Zhang and Russ Salakhutdinov and Ryan Andrew Chi and Seungjae Ryan Lee and Ryan Stovall and Ryan Teehan and Rylan Yang and Sahib Singh and Saif M. Mohammad and Sajant Anand and Sam Dillavou and Sam Shleifer and Sam Wiseman and Samuel Gruetter and Samuel R. Bowman and Samuel Stern Schoenholz and Sanghyun Han and Sanjeev Kwatra and Sarah A. Rous and Sarik Ghazarian and Sayan Ghosh and Sean Casey and Sebastian Bischoff and Sebastian Gehrmann and Sebastian Schuster and Sepideh Sadeghi and Shadi Hamdan and Sharon Zhou and Shashank Srivastava and Sherry Shi and Shikhar Singh and Shima Asaadi and Shixiang Shane Gu and Shubh Pachchigar and Shubham Toshniwal and Shyam Upadhyay and Shyamolima Shammie Debnath and Siamak Shakeri and Simon Thormeyer and Simone Melzi and Siva Reddy and Sneha Priscilla Makini and Soo-Hwan Lee and Spencer Torene and Sriharsha Hatwar and Stanislas Dehaene and Stefan Divic and Stefano Ermon and Stella Biderman and Stephanie Lin and Stephen Prasad and Steven Piantadosi and Stuart Shieber and Summer Misherghi and Svetlana Kiritchenko and Swaroop Mishra and Tal Linzen and Tal Schuster and Tao Li and Tao Yu and Tariq Ali and Tatsunori Hashimoto and Te-Lin Wu and Th{\'e}o Desbordes and Theodore Rothschild and Thomas Phan and Tianle Wang and Tiberius Nkinyili and Timo Schick and Timofei Kornev and Titus Tunduny and Tobias Gerstenberg and Trenton Chang and Trishala Neeraj and Tushar Khot and Tyler Shultz and Uri Shaham and Vedant Misra and Vera Demberg and Victoria Nyamai and Vikas Raunak and Vinay Venkatesh Ramasesh and vinay uday prabhu and Vishakh Padmakumar and Vivek Srikumar and William Fedus and William Saunders and William Zhang and Wout Vossen and Xiang Ren and Xiaoyu Tong and Xinran Zhao and Xinyi Wu and Xudong Shen and Yadollah Yaghoobzadeh and Yair Lakretz and Yangqiu Song and Yasaman Bahri and Yejin Choi and Yichi Yang and Sophie Hao and Yifu Chen and Yonatan Belinkov and Yu Hou and Yufang Hou and Yuntao Bai and Zachary Seid and Zhuoye Zhao and Zijian Wang and Zijie J. Wang and Zirui Wang and Ziyi Wu},
journal={Transactions on Machine Learning Research},
issn={2835-8856},
year={2023},
url={https://openreview.net/forum?id=uyTL5Bvosj},
note={Featured Certification}
}

@misc{
wang2024tram,
title={{TRAM}: Benchmarking Temporal Reasoning for Large Language Models},
author={Yuqing Wang and Yun Zhao},
year={2024},
url={https://openreview.net/forum?id=EJvFFedM2I}
}

@inproceedings{xiong2024large,
  title={Large language models can learn temporal reasoning},
  author={Xiong, Siheng and Payani, Ali and Kompella, Ramana and Fekri, Faramarz},
  booktitle={Proceedings of the 62nd Annual Meeting of the Association for Computational Linguistics (Volume 1: Long Papers)},
  pages={10452--10470},
  year={2024}
}

@inproceedings{tan-etal-2023-towards,
    title = "Towards Benchmarking and Improving the Temporal Reasoning Capability of Large Language Models",
    author = "Tan, Qingyu  and
      Ng, Hwee Tou  and
      Bing, Lidong",
    editor = "Rogers, Anna  and
      Boyd-Graber, Jordan  and
      Okazaki, Naoaki",
    booktitle = "Proceedings of the 61st Annual Meeting of the Association for Computational Linguistics (Volume 1: Long Papers)",
    month = jul,
    year = "2023",
    address = "Toronto, Canada",
    publisher = "Association for Computational Linguistics",
    url = "https://aclanthology.org/2023.acl-long.828/",
    doi = "10.18653/v1/2023.acl-long.828",
    pages = "14820--14835"
}

@article{wei2022chain,
  title={Chain-of-thought prompting elicits reasoning in large language models},
  author={Wei, Jason and Wang, Xuezhi and Schuurmans, Dale and Bosma, Maarten and Xia, Fei and Chi, Ed and Le, Quoc V and Zhou, Denny and others},
  journal={Advances in neural information processing systems},
  volume={35},
  pages={24824--24837},
  year={2022}
}

@misc{bubeck2023sparksartificialgeneralintelligence,
      title={Sparks of Artificial General Intelligence: Early experiments with GPT-4}, 
      author={Sébastien Bubeck and Varun Chandrasekaran and Ronen Eldan and Johannes Gehrke and Eric Horvitz and Ece Kamar and Peter Lee and Yin Tat Lee and Yuanzhi Li and Scott Lundberg and Harsha Nori and Hamid Palangi and Marco Tulio Ribeiro and Yi Zhang},
      year={2023},
      eprint={2303.12712},
      archivePrefix={arXiv},
      primaryClass={cs.CL},
      url={https://arxiv.org/abs/2303.12712}, 
}

@inproceedings{
wang2023selfconsistency,
title={Self-Consistency Improves Chain of Thought Reasoning in Language Models},
author={Xuezhi Wang and Jason Wei and Dale Schuurmans and Quoc V Le and Ed H. Chi and Sharan Narang and Aakanksha Chowdhery and Denny Zhou},
booktitle={The Eleventh International Conference on Learning Representations },
year={2023},
url={https://openreview.net/forum?id=1PL1NIMMrw}
}

@inproceedings{10.5555/3666122.3666639ToT,
author = {Yao, Shunyu and Yu, Dian and Zhao, Jeffrey and Shafran, Izhak and Griffiths, Thomas L. and Cao, Yuan and Narasimhan, Karthik},
title = {Tree of thoughts: deliberate problem solving with large language models},
year = {2023},
publisher = {Curran Associates Inc.},
address = {Red Hook, NY, USA},
booktitle = {Proceedings of the 37th International Conference on Neural Information Processing Systems},
articleno = {517},
numpages = {14},
location = {New Orleans, LA, USA},
series = {NIPS '23}
}

@inproceedings{10.1609/aaai.v38i16.29720GOT,
author = {Besta, Maciej and Blach, Nils and Kubicek, Ales and Gerstenberger, Robert and Podstawski, Micha\l{} and Gianinazzi, Lukas and Gajda, Joanna and Lehmann, Tomasz and Niewiadomski, Hubert and Nyczyk, Piotr and Hoefler, Torsten},
title = {Graph of thoughts: solving elaborate problems with large language models},
year = {2024},
isbn = {978-1-57735-887-9},
publisher = {AAAI Press},
url = {https://doi.org/10.1609/aaai.v38i16.29720},
doi = {10.1609/aaai.v38i16.29720},
booktitle = {Proceedings of the Thirty-Eighth AAAI Conference on Artificial Intelligence and Thirty-Sixth Conference on Innovative Applications of Artificial Intelligence and Fourteenth Symposium on Educational Advances in Artificial Intelligence},
articleno = {1972},
numpages = {9},
series = {AAAI'24/IAAI'24/EAAI'24}
}

@article{
chen2023program,
title={Program of Thoughts Prompting: Disentangling Computation from Reasoning for Numerical Reasoning Tasks},
author={Wenhu Chen and Xueguang Ma and Xinyi Wang and William W. Cohen},
journal={Transactions on Machine Learning Research},
issn={2835-8856},
year={2023},
url={https://openreview.net/forum?id=YfZ4ZPt8zd},
note={}
}

@article{JMLR:v25:22-1258,
author  = {Patrick Bl{{\"o}}baum and Peter G{{\"o}}tz and Kailash Budhathoki and Atalanti A. Mastakouri and Dominik Janzing},
title   = {DoWhy-GCM: An Extension of DoWhy for Causal Inference in Graphical Causal Models},
journal = {Journal of Machine Learning Research},
year    = {2024},
volume  = {25},
number  = {147},
pages   = {1--7},
url     = {http://jmlr.org/papers/v25/22-1258.html}
}

@inproceedings{wei2025satbench,
  title={Satbench: Benchmarking llms’ logical reasoning via automated puzzle generation from sat formulas},
  author={Wei, Anjiang and Wu, Yuheng and Wan, Yingjia and Suresh, Tarun and Tan, Huanmi and Zhou, Zhanke and Koyejo, Sanmi and Wang, Ke and Aiken, Alex},
  booktitle={Proceedings of the 2025 Conference on Empirical Methods in Natural Language Processing},
  pages={33820--33837},
  year={2025}
}

@misc{mukherjee2023orcaprogressivelearningcomplex,
      title={Orca: Progressive Learning from Complex Explanation Traces of GPT-4}, 
      author={Subhabrata Mukherjee and Arindam Mitra and Ganesh Jawahar and Sahaj Agarwal and Hamid Palangi and Ahmed Awadallah},
      year={2023},
      eprint={2306.02707},
      archivePrefix={arXiv},
      primaryClass={cs.CL},
      url={https://arxiv.org/abs/2306.02707}, 
}

@inproceedings{
toshniwal2024openmathinstruct,
title={OpenMathInstruct-1: A 1.8 Million Math Instruction Tuning Dataset},
author={Shubham Toshniwal and Ivan Moshkov and Sean Narenthiran and Daria Gitman and Fei Jia and Igor Gitman},
booktitle={The Thirty-eight Conference on Neural Information Processing Systems Datasets and Benchmarks Track},
year={2024},
url={https://openreview.net/forum?id=Mbd3QxXjq5}
}

@inproceedings{huang2025opencoder,
  title={Opencoder: The open cookbook for top-tier code large language models},
  author={Huang, Siming and Cheng, Tianhao and Liu, Jason Klein and Xu, Weidi and Hao, Jiaran and Song, Liuyihan and Xu, Yang and Yang, Jian and Liu, Jiaheng and Zhang, Chenchen and others},
  booktitle={Proceedings of the 63rd Annual Meeting of the Association for Computational Linguistics (Volume 1: Long Papers)},
  pages={33167--33193},
  year={2025}
}

@misc{qwen2025qwen25technicalreport,
      title={Qwen2.5 Technical Report}, 
      author={Qwen and : and An Yang and Baosong Yang and Beichen Zhang and Binyuan Hui and Bo Zheng and Bowen Yu and Chengyuan Li and Dayiheng Liu and Fei Huang and Haoran Wei and Huan Lin and Jian Yang and Jianhong Tu and Jianwei Zhang and Jianxin Yang and Jiaxi Yang and Jingren Zhou and Junyang Lin and Kai Dang and Keming Lu and Keqin Bao and Kexin Yang and Le Yu and Mei Li and Mingfeng Xue and Pei Zhang and Qin Zhu and Rui Men and Runji Lin and Tianhao Li and Tianyi Tang and Tingyu Xia and Xingzhang Ren and Xuancheng Ren and Yang Fan and Yang Su and Yichang Zhang and Yu Wan and Yuqiong Liu and Zeyu Cui and Zhenru Zhang and Zihan Qiu},
      year={2025},
      eprint={2412.15115},
      archivePrefix={arXiv},
      primaryClass={cs.CL},
      url={https://arxiv.org/abs/2412.15115}, 
}

@article{sakaguchi2021winogrande,
  title={Winogrande: An adversarial winograd schema challenge at scale},
  author={Sakaguchi, Keisuke and Bras, Ronan Le and Bhagavatula, Chandra and Choi, Yejin},
  journal={Communications of the ACM},
  volume={64},
  number={9},
  pages={99--106},
  year={2021},
  publisher={ACM New York, NY, USA}
}

@inproceedings{suzgun2023challenging,
  title={Challenging big-bench tasks and whether chain-of-thought can solve them},
  author={Suzgun, Mirac and Scales, Nathan and Sch{\"a}rli, Nathanael and Gehrmann, Sebastian and Tay, Yi and Chung, Hyung Won and Chowdhery, Aakanksha and Le, Quoc and Chi, Ed and Zhou, Denny and others},
  booktitle={Findings of the Association for Computational Linguistics: ACL 2023},
  pages={13003--13051},
  year={2023}
}

@article{zhou2023instruction,
  title={Instruction-following evaluation for large language models},
  author={Zhou, Jeffrey and Lu, Tianjian and Mishra, Swaroop and Brahma, Siddhartha and Basu, Sujoy and Luan, Yi and Zhou, Denny and Hou, Le},
  journal={arXiv preprint arXiv:2311.07911},
  year={2023}
}

@article{lewkowycz2022solving,
  title={Solving quantitative reasoning problems with language models},
  author={Lewkowycz, Aitor and Andreassen, Anders and Dohan, David and Dyer, Ethan and Michalewski, Henryk and Ramasesh, Vinay and Slone, Ambrose and Anil, Cem and Schlag, Imanol and Gutman-Solo, Theo and others},
  journal={Advances in neural information processing systems},
  volume={35},
  pages={3843--3857},
  year={2022}
}

@article{clark2018think,
  title={Think you have solved question answering? try arc, the ai2 reasoning challenge},
  author={Clark, Peter and Cowhey, Isaac and Etzioni, Oren and Khot, Tushar and Sabharwal, Ashish and Schoenick, Carissa and Tafjord, Oyvind},
  journal={arXiv preprint arXiv:1803.05457},
  year={2018}
}

@inproceedings{
hendrycks2021measuring,
title={Measuring Massive Multitask Language Understanding},
author={Dan Hendrycks and Collin Burns and Steven Basart and Andy Zou and Mantas Mazeika and Dawn Song and Jacob Steinhardt},
booktitle={International Conference on Learning Representations},
year={2021},
url={https://openreview.net/forum?id=d7KBjmI3GmQ}
}

@inproceedings{zellers2019hellaswag,
  title={Hellaswag: Can a machine really finish your sentence?},
  author={Zellers, Rowan and Holtzman, Ari and Bisk, Yonatan and Farhadi, Ali and Choi, Yejin},
  booktitle={Proceedings of the 57th annual meeting of the association for computational linguistics},
  pages={4791--4800},
  year={2019}
}

@misc{eval-harness,
  author       = {Gao, Leo and Tow, Jonathan and Abbasi, Baber and Biderman, Stella and Black, Sid and DiPofi, Anthony and Foster, Charles and Golding, Laurence and Hsu, Jeffrey and Le Noac'h, Alain and Li, Haonan and McDonell, Kyle and Muennighoff, Niklas and Ociepa, Chris and Phang, Jason and Reynolds, Laria and Schoelkopf, Hailey and Skowron, Aviya and Sutawika, Lintang and Tang, Eric and Thite, Anish and Wang, Ben and Wang, Kevin and Zou, Andy},
  title        = {The Language Model Evaluation Harness},
  month        = 07,
  year         = 2024,
  publisher    = {Zenodo},
  version      = {v0.4.3},
  doi          = {10.5281/zenodo.12608602},
  url          = {https://zenodo.org/records/12608602}
}

@article{halpern2005causes,
  title={Causes and explanations: A structural-model approach. Part I: Causes},
  author={Halpern, Joseph Y and Pearl, Judea},
  journal={The British journal for the philosophy of science},
  year={2005},
  publisher={The University of Chicago Press}
}

@book{pearl2009causality,
  title={Causality},
  author={Pearl, Judea},
  year={2009},
  publisher={Cambridge university press}
}

@book{halpern2016actual,
  title={Actual causality},
  author={Halpern, Joseph Y},
  year={2016},
  publisher={MiT Press}
}

@misc{grattafiori_llama_2024,
    title = {The {Llama} 3 {Herd} of {Models}},
    url = {http://arxiv.org/abs/2407.21783},
    doi = {10.48550/arXiv.2407.21783},
    language = {en},
    urldate = {2025-05-07},
    publisher = {arXiv},
    author = {Grattafiori, Aaron and Dubey, Abhimanyu and Jauhri, Abhinav and Pandey, Abhinav and Kadian, Abhishek and Al-Dahle, Ahmad and Letman, Aiesha and Mathur, Akhil and Schelten, Alan and Vaughan, Alex and Yang, Amy and Fan, Angela and Goyal, Anirudh and Hartshorn, Anthony and Yang, Aobo and Mitra, Archi and Sravankumar, Archie and Korenev, Artem and Hinsvark, Arthur and Rao, Arun and Zhang, Aston and Rodriguez, Aurelien and Gregerson, Austen and Spataru, Ava and Roziere, Baptiste and Biron, Bethany and Tang, Binh and Chern, Bobbie and Caucheteux, Charlotte and Nayak, Chaya and Bi, Chloe and Marra, Chris and McConnell, Chris and Keller, Christian and Touret, Christophe and Wu, Chunyang and Wong, Corinne and Ferrer, Cristian Canton and Nikolaidis, Cyrus and Allonsius, Damien and Song, Daniel and Pintz, Danielle and Livshits, Danny and Wyatt, Danny and Esiobu, David and Choudhary, Dhruv and Mahajan, Dhruv and Garcia-Olano, Diego and Perino, Diego and Hupkes, Dieuwke and Lakomkin, Egor and AlBadawy, Ehab and Lobanova, Elina and Dinan, Emily and Smith, Eric Michael and Radenovic, Filip and Guzmán, Francisco and Zhang, Frank and Synnaeve, Gabriel and Lee, Gabrielle and Anderson, Georgia Lewis and Thattai, Govind and Nail, Graeme and Mialon, Gregoire and Pang, Guan and Cucurell, Guillem and Nguyen, Hailey and Korevaar, Hannah and Xu, Hu and Touvron, Hugo and Zarov, Iliyan and Ibarra, Imanol Arrieta and Kloumann, Isabel and Misra, Ishan and Evtimov, Ivan and Zhang, Jack and Copet, Jade and Lee, Jaewon and Geffert, Jan and Vranes, Jana and Park, Jason and Mahadeokar, Jay and Shah, Jeet and Linde, Jelmer van der and Billock, Jennifer and Hong, Jenny and Lee, Jenya and Fu, Jeremy and Chi, Jianfeng and Huang, Jianyu and Liu, Jiawen and Wang, Jie and Yu, Jiecao and Bitton, Joanna and Spisak, Joe and Park, Jongsoo and Rocca, Joseph and Johnstun, Joshua and Saxe, Joshua and Jia, Junteng and Alwala, Kalyan Vasuden and Prasad, Karthik and Upasani, Kartikeya and Plawiak, Kate and Li, Ke and Heafield, Kenneth and Stone, Kevin and El-Arini, Khalid and Iyer, Krithika and Malik, Kshitiz and Chiu, Kuenley and Bhalla, Kunal and Lakhotia, Kushal and Rantala-Yeary, Lauren and Maaten, Laurens van der and Chen, Lawrence and Tan, Liang and Jenkins, Liz and Martin, Louis and Madaan, Lovish and Malo, Lubo and Blecher, Lukas and Landzaat, Lukas and Oliveira, Luke de and Muzzi, Madeline and Pasupuleti, Mahesh and Singh, Mannat and Paluri, Manohar and Kardas, Marcin and Tsimpoukelli, Maria and Oldham, Mathew and Rita, Mathieu and Pavlova, Maya and Kambadur, Melanie and Lewis, Mike and Si, Min and Singh, Mitesh Kumar and Hassan, Mona and Goyal, Naman and Torabi, Narjes and Bashlykov, Nikolay and Bogoychev, Nikolay and Chatterji, Niladri and Zhang, Ning and Duchenne, Olivier and Çelebi, Onur and Alrassy, Patrick and Zhang, Pengchuan and Li, Pengwei and Vasic, Petar and Weng, Peter and Bhargava, Prajjwal and Dubal, Pratik and Krishnan, Praveen and Koura, Punit Singh and Xu, Puxin and He, Qing and Dong, Qingxiao and Srinivasan, Ragavan and Ganapathy, Raj and Calderer, Ramon and Cabral, Ricardo Silveira and Stojnic, Robert and Raileanu, Roberta and Maheswari, Rohan and Girdhar, Rohit and Patel, Rohit and Sauvestre, Romain and Polidoro, Ronnie and Sumbaly, Roshan and Taylor, Ross and Silva, Ruan and Hou, Rui and Wang, Rui and Hosseini, Saghar and Chennabasappa, Sahana and Singh, Sanjay and Bell, Sean and Kim, Seohyun Sonia and Edunov, Sergey and Nie, Shaoliang and Narang, Sharan and Raparthy, Sharath and Shen, Sheng and Wan, Shengye and Bhosale, Shruti and Zhang, Shun and Vandenhende, Simon and Batra, Soumya and Whitman, Spencer and Sootla, Sten and Collot, Stephane and Gururangan, Suchin and Borodinsky, Sydney and Herman, Tamar and Fowler, Tara and Sheasha, Tarek and Georgiou, Thomas and Scialom, Thomas and Speckbacher, Tobias and Mihaylov, Todor and Xiao, Tong and Karn, Ujjwal and Goswami, Vedanuj and Gupta, Vibhor and Ramanathan, Vignesh and Kerkez, Viktor and Gonguet, Vincent and Do, Virginie and Vogeti, Vish and Albiero, Vítor and Petrovic, Vladan and Chu, Weiwei and Xiong, Wenhan and Fu, Wenyin and Meers, Whitney and Martinet, Xavier and Wang, Xiaodong and Wang, Xiaofang and Tan, Xiaoqing Ellen and Xia, Xide and Xie, Xinfeng and Jia, Xuchao and Wang, Xuewei and Goldschlag, Yaelle and Gaur, Yashesh and Babaei, Yasmine and Wen, Yi and Song, Yiwen and Zhang, Yuchen and Li, Yue and Mao, Yuning and Coudert, Zacharie Delpierre and Yan, Zheng and Chen, Zhengxing and Papakipos, Zoe and Singh, Aaditya and Srivastava, Aayushi and Jain, Abha and Kelsey, Adam and Shajnfeld, Adam and Gangidi, Adithya and Victoria, Adolfo and Goldstand, Ahuva and Menon, Ajay and Sharma, Ajay and Boesenberg, Alex and Baevski, Alexei and Feinstein, Allie and Kallet, Amanda and Sangani, Amit and Teo, Amos and Yunus, Anam and Lupu, Andrei and Alvarado, Andres and Caples, Andrew and Gu, Andrew and Ho, Andrew and Poulton, Andrew and Ryan, Andrew and Ramchandani, Ankit and Dong, Annie and Franco, Annie and Goyal, Anuj and Saraf, Aparajita and Chowdhury, Arkabandhu and Gabriel, Ashley and Bharambe, Ashwin and Eisenman, Assaf and Yazdan, Azadeh and James, Beau and Maurer, Ben and Leonhardi, Benjamin and Huang, Bernie and Loyd, Beth and Paola, Beto De and Paranjape, Bhargavi and Liu, Bing and Wu, Bo and Ni, Boyu and Hancock, Braden and Wasti, Bram and Spence, Brandon and Stojkovic, Brani and Gamido, Brian and Montalvo, Britt and Parker, Carl and Burton, Carly and Mejia, Catalina and Liu, Ce and Wang, Changhan and Kim, Changkyu and Zhou, Chao and Hu, Chester and Chu, Ching-Hsiang and Cai, Chris and Tindal, Chris and Feichtenhofer, Christoph and Gao, Cynthia and Civin, Damon and Beaty, Dana and Kreymer, Daniel and Li, Daniel and Adkins, David and Xu, David and Testuggine, Davide and David, Delia and Parikh, Devi and Liskovich, Diana and Foss, Didem and Wang, Dingkang and Le, Duc and Holland, Dustin and Dowling, Edward and Jamil, Eissa and Montgomery, Elaine and Presani, Eleonora and Hahn, Emily and Wood, Emily and Le, Eric-Tuan and Brinkman, Erik and Arcaute, Esteban and Dunbar, Evan and Smothers, Evan and Sun, Fei and Kreuk, Felix and Tian, Feng and Kokkinos, Filippos and Ozgenel, Firat and Caggioni, Francesco and Kanayet, Frank and Seide, Frank and Florez, Gabriela Medina and Schwarz, Gabriella and Badeer, Gada and Swee, Georgia and Halpern, Gil and Herman, Grant and Sizov, Grigory and Guangyi and Zhang and Lakshminarayanan, Guna and Inan, Hakan and Shojanazeri, Hamid and Zou, Han and Wang, Hannah and Zha, Hanwen and Habeeb, Haroun and Rudolph, Harrison and Suk, Helen and Aspegren, Henry and Goldman, Hunter and Zhan, Hongyuan and Damlaj, Ibrahim and Molybog, Igor and Tufanov, Igor and Leontiadis, Ilias and Veliche, Irina-Elena and Gat, Itai and Weissman, Jake and Geboski, James and Kohli, James and Lam, Janice and Asher, Japhet and Gaya, Jean-Baptiste and Marcus, Jeff and Tang, Jeff and Chan, Jennifer and Zhen, Jenny and Reizenstein, Jeremy and Teboul, Jeremy and Zhong, Jessica and Jin, Jian and Yang, Jingyi and Cummings, Joe and Carvill, Jon and Shepard, Jon and McPhie, Jonathan and Torres, Jonathan and Ginsburg, Josh and Wang, Junjie and Wu, Kai and U, Kam Hou and Saxena, Karan and Khandelwal, Kartikay and Zand, Katayoun and Matosich, Kathy and Veeraraghavan, Kaushik and Michelena, Kelly and Li, Keqian and Jagadeesh, Kiran and Huang, Kun and Chawla, Kunal and Huang, Kyle and Chen, Lailin and Garg, Lakshya and A, Lavender and Silva, Leandro and Bell, Lee and Zhang, Lei and Guo, Liangpeng and Yu, Licheng and Moshkovich, Liron and Wehrstedt, Luca and Khabsa, Madian and Avalani, Manav and Bhatt, Manish and Mankus, Martynas and Hasson, Matan and Lennie, Matthew and Reso, Matthias and Groshev, Maxim and Naumov, Maxim and Lathi, Maya and Keneally, Meghan and Liu, Miao and Seltzer, Michael L. and Valko, Michal and Restrepo, Michelle and Patel, Mihir and Vyatskov, Mik and Samvelyan, Mikayel and Clark, Mike and Macey, Mike and Wang, Mike and Hermoso, Miquel Jubert and Metanat, Mo and Rastegari, Mohammad and Bansal, Munish and Santhanam, Nandhini and Parks, Natascha and White, Natasha and Bawa, Navyata and Singhal, Nayan and Egebo, Nick and Usunier, Nicolas and Mehta, Nikhil and Laptev, Nikolay Pavlovich and Dong, Ning and Cheng, Norman and Chernoguz, Oleg and Hart, Olivia and Salpekar, Omkar and Kalinli, Ozlem and Kent, Parkin and Parekh, Parth and Saab, Paul and Balaji, Pavan and Rittner, Pedro and Bontrager, Philip and Roux, Pierre and Dollar, Piotr and Zvyagina, Polina and Ratanchandani, Prashant and Yuvraj, Pritish and Liang, Qian and Alao, Rachad and Rodriguez, Rachel and Ayub, Rafi and Murthy, Raghotham and Nayani, Raghu and Mitra, Rahul and Parthasarathy, Rangaprabhu and Li, Raymond and Hogan, Rebekkah and Battey, Robin and Wang, Rocky and Howes, Russ and Rinott, Ruty and Mehta, Sachin and Siby, Sachin and Bondu, Sai Jayesh and Datta, Samyak and Chugh, Sara and Hunt, Sara and Dhillon, Sargun and Sidorov, Sasha and Pan, Satadru and Mahajan, Saurabh and Verma, Saurabh and Yamamoto, Seiji and Ramaswamy, Sharadh and Lindsay, Shaun and Lindsay, Shaun and Feng, Sheng and Lin, Shenghao and Zha, Shengxin Cindy and Patil, Shishir and Shankar, Shiva and Zhang, Shuqiang and Zhang, Shuqiang and Wang, Sinong and Agarwal, Sneha and Sajuyigbe, Soji and Chintala, Soumith and Max, Stephanie and Chen, Stephen and Kehoe, Steve and Satterfield, Steve and Govindaprasad, Sudarshan and Gupta, Sumit and Deng, Summer and Cho, Sungmin and Virk, Sunny and Subramanian, Suraj and Choudhury, Sy and Goldman, Sydney and Remez, Tal and Glaser, Tamar and Best, Tamara and Koehler, Thilo and Robinson, Thomas and Li, Tianhe and Zhang, Tianjun and Matthews, Tim and Chou, Timothy and Shaked, Tzook and Vontimitta, Varun and Ajayi, Victoria and Montanez, Victoria and Mohan, Vijai and Kumar, Vinay Satish and Mangla, Vishal and Ionescu, Vlad and Poenaru, Vlad and Mihailescu, Vlad Tiberiu and Ivanov, Vladimir and Li, Wei and Wang, Wenchen and Jiang, Wenwen and Bouaziz, Wes and Constable, Will and Tang, Xiaocheng and Wu, Xiaojian and Wang, Xiaolan and Wu, Xilun and Gao, Xinbo and Kleinman, Yaniv and Chen, Yanjun and Hu, Ye and Jia, Ye and Qi, Ye and Li, Yenda and Zhang, Yilin and Zhang, Ying and Adi, Yossi and Nam, Youngjin and Yu and Wang and Zhao, Yu and Hao, Yuchen and Qian, Yundi and Li, Yunlu and He, Yuzi and Rait, Zach and DeVito, Zachary and Rosnbrick, Zef and Wen, Zhaoduo and Yang, Zhenyu and Zhao, Zhiwei and Ma, Zhiyu},
    month = nov,
    year = {2024},
    note = {arXiv:2407.21783 [cs]},
}

@article{hu2022lora,
  title={Lora: Low-rank adaptation of large language models.},
  author={Hu, Edward J and Shen, Yelong and Wallis, Phillip and Allen-Zhu, Zeyuan and Li, Yuanzhi and Wang, Shean and Wang, Liang and Chen, Weizhu and others},
  journal={Iclr},
  volume={1},
  number={2},
  pages={3},
  year={2022}
}
\bibliographystyle{colm2026_conference}

\newpage
\appendix
\onecolumn

\section{Details of SYNTCOMP Data Used}
\label{app:syntcomp}
\textsc{TempoBench} draws its evaluation and training automata from the SYNTCOMP benchmark suite~\cite{syntcomp_benchmarks}, a standardized collection of reactive synthesis problems specified in TLSF (Temporal Logic Synthesis Format)~\cite{DBLP:journals/corr/Jacobs016}. We use specifications from three families:

\textbf{Arbiters} (\texttt{arbiter\_with\_buffer\_prio\_param\_n2}, \texttt{n3}): Model mutual exclusion protocols where $n$ processes compete for shared resources. The synthesized controllers enforce fairness and priority constraints. These produce automata with 7--17 states and 14--72 transitions.

\textbf{Chomp games} (\texttt{chomp\_n2m2}, \texttt{n3m3}): Model the combinatorial game Chomp on an $n \times m$ grid, where the controller must find a winning strategy. These produce larger automata (up to 69 states, 707 transitions) with complex branching.

\textbf{Load balancers} (\texttt{load\_balancer\_n2}, \texttt{n3}, \texttt{n4}): Model request distribution across $n$ servers with fairness guarantees. These produce moderately sized automata (8--30 states) with dense transition tables.

Across all specifications, automata range from 7 to 69 states (mean 11.5) and 14 to 707 transitions (mean 38.6). The evaluation set comprises 1,000 samples (500 SIM + 500 MIN) drawn from 10 specification types. Trace lengths range from 3 to 15 steps, with 505 easy samples (path length $\leq 6$) and 495 hard samples (path length $> 6$). Training data is generated deterministically from the same pipeline, train and test sets are drawn from disjoint traces to prevent contamination.

\section{Details on Selected Benchmarks}
\label{app:benchmarks}
We evaluate fine-tuned models on eight standard benchmarks spanning mathematical reasoning, language understanding, instruction following, and general knowledge:

\textbf{ARC-Challenge}~\cite{clark2018think}: Grade-school science questions requiring multi-step reasoning. We report normalized accuracy (acc\_norm).

\textbf{WinoGrande}~\cite{sakaguchi2021winogrande}: Commonsense pronoun resolution at scale. We report accuracy.

\textbf{GSM8K}~\cite{cobbe2021trainingverifierssolvemath}: 8,500 grade-school math word problems requiring multi-step arithmetic. We report exact match with strict matching.

\textbf{MMLU}~\cite{hendrycks2021measuring}: Massive Multitask Language Understanding covering 57 subjects. We report 5-shot accuracy.

\textbf{HellaSwag}~\cite{zellers2019hellaswag}: Commonsense natural language inference requiring grounded physical reasoning. We report normalized accuracy.

\textbf{MINERVA Math}~\cite{lewkowycz2022solving}: Challenging mathematical reasoning problems from STEM courses. We report exact match; this benchmark is harder than GSM8K and tests for generalization vs.\ overfitting.

\textbf{IFEval}~\cite{zhou2023instruction}: Instruction-following evaluation measuring strict compliance with explicit formatting constraints. We report instruction-level strict accuracy.

\textbf{BBH-CoT}~\cite{suzgun2023challenging}: BIG-Bench Hard with chain-of-thought prompting, covering 23 challenging tasks. We report exact match with the get-answer extraction.

All evaluations use the \texttt{lm\_eval} harness~\cite{eval-harness} with default settings per task.

\section{Training Dataset Comparison}
\label{app:dataset-comparison}
\begin{table}[h]
  \centering
  \small
  \setlength{\tabcolsep}{4pt}
  \begin{tabular}{lrrrr}
    \toprule
    & \textbf{TB} & \textbf{SO} & \textbf{OMI-1} & \textbf{OC} \\
    \midrule
    Mean input tokens   & 2{,}615 & 252 & 56 & 262 \\
    Mean output tokens  & 2{,}963 & 130 & 162 & 233 \\
    Mean total tokens   & 5{,}578 & 382 & 218 & 495 \\
    Output / total      & 53\% & 34\% & 74\% & 47\% \\
    Dataset total (M tokens) & 278.9 & 19.1 & 10.9 & 24.8 \\
    \bottomrule
  \end{tabular}
  \caption{\textbf{Token-level statistics for 50k-sample training sets.} \textsc{TempoBench} samples are 10--25$\times$ longer than other datasets because each includes the full Mealy machine description and a chain-of-thought reasoning trace. Tokenized with cl100k\_base.}
  \label{tab:dataset-stats}
\end{table}

The key difference is the nature of the training signal. \textsc{TempoBench} samples are supervised primarily on the reasoning trace (CoT), with only a small fraction supervised on the final structured answer (Section~\ref{sec:benchmark}). The model learns to reason step-by-step through the automaton rather than to produce answer patterns. In contrast, SO provides short instruction--response pairs (mean 382 tokens), OMI-1 provides concise math solutions (mean 218 tokens, 74\% output), and OC provides code completions (mean 495 tokens). Our pipeline generates nearly 300M supervision tokens within hours on consumer hardware. 

\section{Structural Factors Influencing Complexity of Temporal Reasoning}
\label{app:complexity-sweep}
\label{sec:analysis-complexity}

\begin{figure*}[h]
    \centering
    \includegraphics[width=\textwidth]{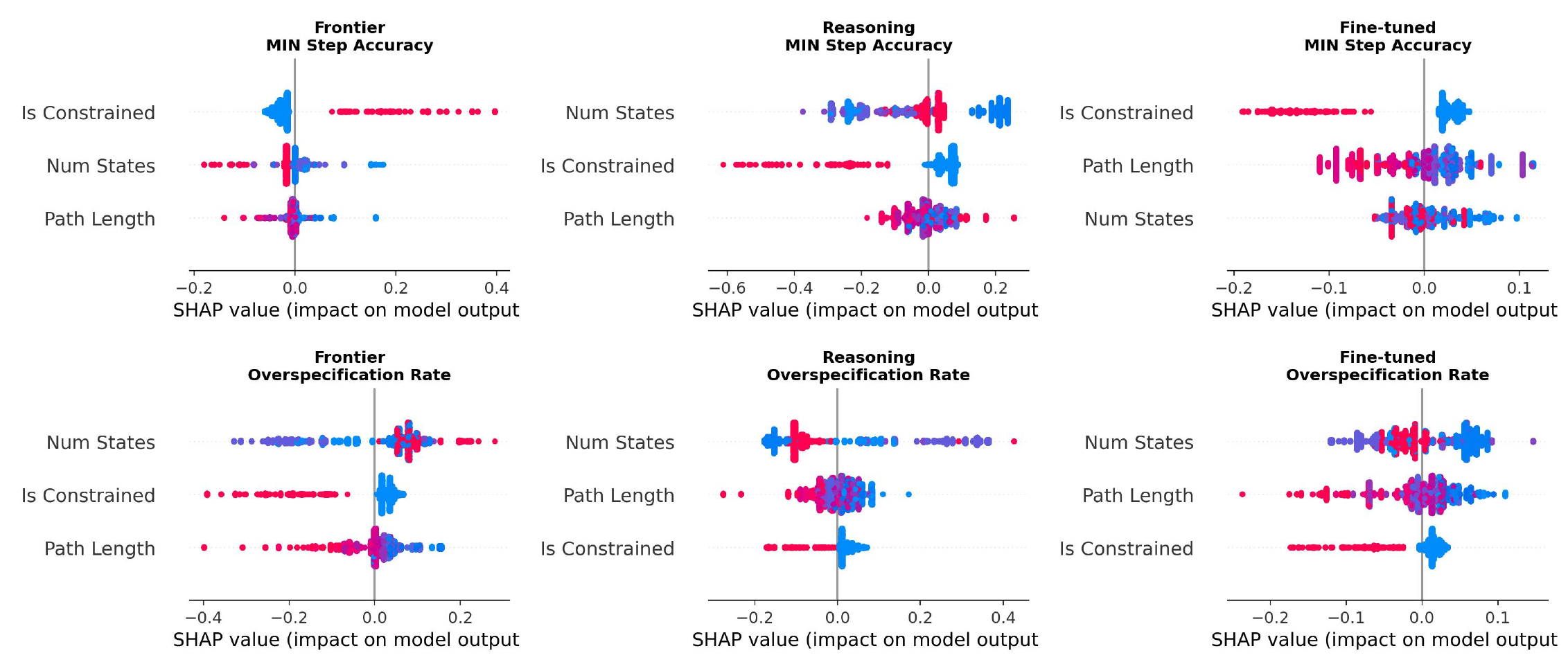}
    \vspace{-10pt}
    \caption{\textbf{SHAP feature importance for MIN.} Top: impact on causal step accuracy. Bottom: impact on overspecification rate. Features are sorted by their impact. Within each feature, each dot is one evaluation sample; color indicates feature value (red = high, blue = low); horizontal position shows the SHAP value (positive = increases the predicted metric). Computed via gradient boosting regression.}
    \vspace{-10pt}
    \label{fig:feature-importance}
\end{figure*}

We fit gradient boosting regressors on all frontier evaluations and compute SHAP values to quantify each structural feature's contribution to MIN step accuracy and overspecification rate (Figure~\ref{fig:feature-importance}). We see frontier models do better when fewer steps are unconstrained; this result supports the paper's claim that LLMs struggle to \textit{prune} noisy information from the execution of a system when doing root cause analysis style tasks. We observe a flipped behavior in the finetuned models; this is likely due to the models learning to exploit this unconstrained gap on the MIN task, the MIN+ and MIN-Hard tasks address this issue. The reasoning models are more resistant to the no constraints condition; instead, they are more sensitive to the number of states, indicating the models were attempting to fully simulate execution (around 20k tokens used per evaluation sample). We observe that fewer states increase overspecification (likely because smaller machines have denser transition tables with more guard literals). Path length has a moderate effect; trace length and effect depth have minimal impact. This indicates that models are bottlenecked by structural errors in reasoning, not by task complexity. Notably, SIM accuracy is nearly independent of all features (not shown), confirming that trace execution is solved by retrieval regardless of machine structure.

\section{Full Error Breakdown (Parsed Steps Only)}
\label{app:parse-success}
\begin{table*}[h]
  \centering
  \setlength{\tabcolsep}{2.5pt}
  \small
  \resizebox{\textwidth}{!}{%
  \begin{tabular}{l|c ccc|c ccc|c ccc}
    \toprule
    & \multicolumn{4}{c|}{\textbf{MIN (parsed only)}} & \multicolumn{4}{c|}{\textbf{MIN+ (parsed only)}} & \multicolumn{4}{c}{\textbf{MIN-Hard (parsed only)}} \\
    \textbf{Model} & Parse & Correct & G.Copy & Unsp & Parse & Correct & G.Copy & Unsp & Parse & Correct & G.Copy & Unsp \\
    \midrule
    \multicolumn{13}{l}{\textit{Frontier models (zero-shot)}} \\
    \cmidrule(l){1-13}
    Sonnet 4.6     & 97.4 & 36.1 & 49.1 &  3.2 & 97.4 & 18.6 & 28.1 &  7.7 & 99.8 &  5.4 & 10.4 & 13.8 \\
    Haiku 4.5      & 100.0 &  4.0 & 93.7 &  0.2 & 100.0 & 24.4 & 58.5 &  1.7 & 100.0 &  3.7 & 93.6 &  0.5 \\
    Gemini Pro     & 13.3 & 56.2 &  4.7 & 16.3 & 13.3 & 28.6 &  1.8 & 44.6 & 10.8 &  0.0 &  0.0 & 88.6 \\
    DeepSeek V3.2  & 99.8 &  3.3 & 94.5 &  0.1 & 99.8 & 24.2 & 58.6 &  0.4 & 98.8 &  2.4 & 94.8 &  0.1 \\
    DeepSeek V3    & 53.0 & 19.5 & 60.6 &  2.8 & 53.0 &  8.6 & 22.7 &  9.0 & 86.6 &  2.3 &  1.2 & 13.5 \\
    \cmidrule(l){1-13}
    \multicolumn{13}{l}{\textit{Reasoning models (extended CoT)}} \\
    \cmidrule(l){1-13}
    DeepSeek R1    & 97.2 & 68.5 & 24.4 &  6.0 & 97.2 & 29.7 & 17.2 & 44.5 & 96.8 & 62.7 & 30.6 &  5.9 \\
    V3.2 (reason)  & 77.8 & 75.5 & 16.0 &  7.4 & 77.8 & 27.1 & 11.8 & 53.4 & 75.6 & 72.9 & 19.2 &  7.2 \\
    \midrule
    \multicolumn{13}{l}{\textit{LLaMA 3.1 8B}} \\
    \cmidrule(l){1-13}
    baseline       &  0.3 &  0.0 & 100.0 & 0.0 & 0.0 & 0.0 & 0.0 & 0.0 & 13.9 & 0.0 & 0.0 & 66.2 \\
    SO           & 100.0 &  5.3 & 81.5 &  0.6 & 100.0 &  8.4 & 40.3 &  2.8 & 91.0 & 0.0 & 6.5 & 4.3 \\
    OMI-1           & 80.1 &  0.5 & 87.3 &  0.0 & 80.1 &  0.0 & 41.6 &  0.0 &  0.3 & 0.0 & 0.0 & 25.0 \\
    OC           & 100.0 &  0.4 & 85.2 &  0.0 & 100.0 &  1.4 & 40.3 &  0.0 & 100.0 & 0.0 & 5.9 & 4.9 \\
    OMI-1+SO      & 98.2 &  6.8 & 78.7 &  0.3 & 98.2 &  1.9 & 38.3 &  1.0 & 90.7 & 0.0 & 6.5 & 30.1 \\
    OC+SO      & 100.0 & 16.5 & 71.5 &  3.4 & 100.0 & 15.4 & 33.8 & 15.9 & 100.0 & 0.0 & 5.9 & 32.5 \\
    \cmidrule(l){1-13}
    tb             & 100.0 & 36.3 & 48.1 &  8.1 & 100.0 & 15.6 & 19.8 & 32.6 & 94.0 & 0.0 & 4.8 & 21.6 \\
    tb+SO        & 100.0 & 26.7 & 63.2 &  6.5 & 100.0 & 21.7 & 37.5 & 27.7 & 99.2 & \textbf{12.0} & 2.7 & 49.6 \\
    tb+OMI-1        & 100.0 & 33.7 & 50.8 & 10.1 & 100.0 & 18.4 & 17.9 & 43.1 & 96.7 & 0.4 & 5.0 & 59.3 \\
    tb+OC        & 100.0 & 42.1 & 44.1 &  9.0 & 100.0 & 20.7 & 24.2 & 35.7 & 85.5 & 3.0 & 7.4 & 54.5 \\
    \midrule
    \multicolumn{13}{l}{\textit{LLaMA 3.2 3B}} \\
    \cmidrule(l){1-13}
    baseline       & 0.0 & 0.0 & 0.0 & 0.0 & 0.0 & 0.0 & 0.0 & 0.0 & 15.0 & 0.0 & 0.6 & 66.5 \\
    SO           & 100.0 & 81.8 & 0.0 & 18.2 & 100.0 & 26.8 & 0.0 & 73.2 & 43.2 & 0.0 & 9.6 & 33.1 \\
    OMI-1           & 21.3 & 6.7 & 84.5 & 0.7 & 21.3 & 0.0 & 48.1 & 7.4 & 0.0 & 0.0 & 0.0 & 0.0 \\
    OC           & 55.0 & 36.2 & 50.2 & 8.4 & 55.0 & 15.4 & 21.3 & 41.0 & 87.2 & 0.0 & 5.3 & 5.6 \\
    OMI-1+SO      & 100.0 & 81.8 & 0.0 & 18.2 & 100.0 & 26.8 & 0.0 & 73.2 & 0.3 & 0.0 & 0.0 & 100.0 \\
    OC+SO      & 100.0 & 78.2 & 3.8 & 18.1 & 100.0 & 26.1 & 0.7 & 73.2 & 88.3 & 0.0 & 6.7 & 33.2 \\
    \cmidrule(l){1-13}
    tb             & 98.5 & 25.4 & 62.7 & 3.7 & 98.5 & 19.8 & 37.2 & 14.4 & 98.9 & 1.7 & 5.3 & 47.2 \\
    tb+SO        & 97.4 & 21.3 & 63.7 & 5.1 & 97.4 & 10.0 & 33.0 & 20.4 & 99.2 & 0.0 & 0.0 & 53.8 \\
    tb+OMI-1        & 67.8 & 31.8 & 53.0 & 9.2 & 67.8 & 18.8 & 16.2 & 40.6 & 29.5 & 0.0 & 4.0 & 67.4 \\
    tb+OC        & 97.7 & 14.8 & 69.9 & 3.8 & 97.7 & 10.6 & 28.7 & 15.0 & 12.6 & \textbf{2.1} & 4.2 & 62.9 \\
    \midrule
    \multicolumn{13}{l}{\textit{Qwen 2.5 7B}} \\
    \cmidrule(l){1-13}
    baseline       & 25.1 & 14.1 & 66.8 & 1.0 & 25.1 & 8.8 & 27.2 & 3.2 & 67.8 & 0.0 & 2.8 & 53.8 \\
    SO           & 100.0 & 6.7 & 78.0 & 0.6 & 100.0 & 4.4 & 36.1 & 1.6 & 100.0 & 0.0 & 5.9 & 33.5 \\
    OMI-1           & 100.0 & 2.9 & 82.1 & 0.8 & 100.0 & 9.8 & 30.5 & 4.2 & 99.5 & 0.0 & 4.5 & 23.5 \\
    OC           & 99.4 & 0.2 & 84.5 & 0.1 & 99.4 & 0.9 & 39.2 & 0.7 & 100.0 & 0.0 & 5.8 & 32.9 \\
    OMI-1+SO      & 100.0 & 16.6 & 68.2 & 3.6 & 100.0 & 11.2 & 27.0 & 14.5 & 100.0 & 0.0 & 3.1 & 41.4 \\
    OC+SO      & 100.0 & 13.5 & 70.3 & 5.0 & 100.0 & 4.9 & 30.3 & 24.7 & 100.0 & 0.0 & 5.9 & 34.1 \\
    \cmidrule(l){1-13}
    tb             & 89.2 & 27.8 & 59.9 & 5.9 & 89.2 & 23.2 & 26.5 & 25.9 & 84.2 & 0.0 & 2.8 & 71.2 \\
    tb+SO        & 99.4 & 3.3 & 84.2 & 2.0 & 99.4 & 8.1 & 41.5 & 10.0 & 48.6 & 0.0 & 2.6 & 73.4 \\
    tb+OMI-1        & 84.5 & 27.3 & 58.8 & 5.1 & 84.5 & 12.4 & 34.0 & 18.8 & 80.6 & \textbf{0.1} & 0.1 & 72.1 \\
    tb+OC        & 77.5 & 28.2 & 59.9 & 4.2 & 77.5 & 11.1 & 36.9 & 15.7 & 59.0 & 0.0 & 0.0 & 98.9 \\
    \bottomrule
  \end{tabular}}
  \caption{\textbf{Full error breakdown over parsed steps only.} Parse: \% of samples successfully parsed. Correct/G.Copy/Unsp computed only over parsed steps (denominator excludes parse failures). This separates format compliance from reasoning ability. Compare with Table~\ref{tab:retrieval} which uses the harsh denominator (unparsed = incorrect).}
  \label{tab:parse-success}
\end{table*}

\section{Synthesis Preliminaries}
\label{app:synthesis}
\subsection{Linear Temporal Logic \& Reactive Systems}\label{LTL}
\textit{Note: Throughout the main paper we use ``transition condition'' to refer to what is formally known as a \textbf{guard} in automata theory---the boolean predicate on inputs that must be satisfied for a transition to take place.}

Reactive systems appear across a wide range of domains, from traditional hardware and software to more abstract environments such as biological systems.
Temporal logics, most prominently Linear Temporal Logic (LTL) \cite{pnueli1977temporal}, provide a formal language for specifying the desired behavior of such systems over time.
We consider LTL, which has the following syntax.

\[
  \varphi  ::= \top | a | \varphi \lor \varphi | \neg \varphi
  | X \varphi | \varphi \mathcal{U} \varphi, 
\]

where  $a\in AP$ is an \emph{atomic proposition}, 
$\{\lor,\neg\}$ are the common Boolean operators of \emph{disjunction} and
\emph{negation}, respectively, and
$\{X,\mathcal{U}\}$ are the \emph{next} and \emph{until} temporal
operators, respectively.

Additional temporal operators include  $F$(\emph{finally}), and
 $G$ (\emph{always}), which can be derived from the syntax above.

Building on LTL specifications, reactive synthesis has extended the Church Synthesis problem~\cite{church1963application} to automatically construct implementations specified from LTL specifications~\cite{pnueli1989synthesis}.
This reactive synthesis can be framed as a two-player infinite game between an environment and a system player. 
The set of atomic propositions is partitioned into inputs 
\(I = \{i_{0}, i_{1}, \ldots, i_{n}\}\) (controlled by the environment) 
and outputs 
\(O = \{o_{0}, o_{1}, \ldots, o_{m}\}\) (controlled by the system). 
At each round, the environment first assigns values to all inputs, then the system assigns values to all outputs, producing an infinite trace over 
\(I \cup O\). 
The system wins if it has a strategy to choose outputs such that for every possible sequence of inputs, the resulting infinite trace satisfies the LTL specification \(\varphi\). 
If such a winning strategy exists, the specification is realizable and a corresponding Mealy machine can be extracted. 
The outputted Mealy machine provides a temporal relationship between the atomic propositions \(I\) and  \(O\) that inherits formal guarantees of the specification \(\varphi\). 
\subsection{Temporal Causality}\label{temporal causality}
\begin{algorithm}[h]
\caption{Counterfactual Causal Attribution (per-step)}
\label{alg:counterfactual_causal_attribution}
\begin{algorithmic}[1]
\Require Controller $A$; observed finite trace $\pi = (s_0 \xrightarrow{i_0/o_0} s_1 \cdots s_T)$
\For{$t \in \{0,\dots,T-1\}$}
    \State $\phi_o \gets$ output proposition observed at step $t$ (from $o_t$)
    \State $\mathcal{C}_t \gets \emptyset$ \Comment{necessary causes at time $t$}
    \For{each candidate condition $c$ in $\mathrm{Cand}(s_t,i_t,A)$}
        \State $\pi' \gets \mathrm{Do}(\pi, t, \neg c)$ \Comment{counterfactual intervention at $t$}
        \If{$\pi'$ is feasible under $A$ \textbf{and} $o'_t \not\models \phi_o$}
            \State $\mathcal{C}_t \gets \mathcal{C}_t \cup \{c\}$ \Comment{$c$ is necessary for $\phi_o$}
        \EndIf
    \EndFor
    \State $\mathrm{Cause}_t \gets \mathrm{Minimize}(\mathcal{C}_t)$ \Comment{optional: minimal necessary subset}
\EndFor
\State \Return $\{(t,\phi_o,\mathrm{Cause}_t)\}_{t=0}^{T-1}$
\end{algorithmic}
\end{algorithm}

Algorithm~\ref{alg:counterfactual_causal_attribution} describes the algorithm used to compute the causal sets following Halpern Pearl causality~\cite{halpern2005causes,halpern2016actual} for reactive systems. We use the CORP tool for synthesis~\cite{finkbeiner2024synthesis}.

Causality in temporal systems is more complicated than the analysis of the underlying LTL specification. While LTL specifications describe relationships between inputs and outputs, temporal causality is interested in identifying the minimum causal set over a trace \(\pi\) which describes each of the inputs necessary for some effect \(E\) to occur. Specifically, given a system $\mathcal{T}$ with traces $\pi \in \text{Traces}(\mathcal{T})$, 
let $C \subseteq (2^{I})^{\omega}$ be a cause property over the inputs and 
let $E \subseteq (2^{O})^{\omega}$ be an effect property over the outputs~\cite{coenen2022temporal}. 
We say that $C$ is a cause of $E$ on $\pi$ in $\mathcal{T}$ if three conditions hold:
\begin{enumerate}
    \item $\pi \models C$ and $\pi \models E$
    \item under some counterfactual variation of the inputs, removing or altering $C$ leads to a trace $\pi'$ in which the effect property $E$ no longer holds, which ensures the effect depends on the cause
    \item no strict subset $C' \subset C$ also satisfies these two conditions, guaranteeing that the cause is minimal and not redundant
\end{enumerate} 
Identifying the causal set for some trace is a synthesis problem that identifies, given a Trace \(\pi\) and system \(T\), what inputs at each time step are necessary to produce the desired output. In our credit-assignment work, we focus on outputs at particular time steps. 

The resulting effects take the form $X X \ldots X\, AP^{o_i}$. Our goal is to identify which inputs at each time step \(t_i \in \pi\) produce this effect \(E\), across all traces with this counterfactual structure, and from this synthesize the minimal controller. This controller represents the minimal causal inputs necessary to produce the desired effect at a time step \(t_j\)~\cite{finkbeiner2024synthesis}.

We use an LTL specification of a music app from the syntcomp benchmark to illustrate the difference between simple credit assignment of a single input-output pair and full causal temporal reasoning. The specification synthesizes a controller managing the interplay between user inputs (play/pause button presses, leaving/resuming the app) and system outputs (play/pause commands, internal control state)~\cite{finkbeiner2019temporal} to illustrate the difference between simple credit assignment of a single input-output pair and full causal temporal reasoning.

\begin{figure}[h]
\centering
\begin{tikzpicture}[>=stealth, every node/.style={font=\small},
    state/.style={circle, draw, minimum size=8mm, inner sep=0pt}]
  \node[state] (s1) at (0,0) {$1$};
  \node[state] (s2) at (4,0) {$2$};
  \node[state] (s3) at (8,0) {$3$};
  \node[state, double] (s0) at (12,0) {$0$};

  \draw[->] (-1.2,0) -- (s1);
  \draw[->] (s1) -- node[above, font=\scriptsize] {leave\_app \& pause} (s2);
  \draw[->] (s2) -- node[above, font=\scriptsize] {leave\_app \& !pause} (s3);
  \draw[->] (s3) -- node[above, font=\scriptsize, text width=3cm, align=center]
    {leave\_app \& !music\_playing \& play\_button} (s0);
  \draw[->] (s0) to[loop above, looseness=5] node[above, font=\scriptsize] {$\top$} (s0);
\end{tikzpicture}
\caption{Temporal Causality HOA for Trace of Music Player}
\label{fig:music-player-hoa}
\end{figure}
\begin{example}[Music Player App]\label{example:musicplayer}
Given the specification:
\[
\begin{aligned}
\varphi = 
&\;\Bigl(
    G\bigl(\neg(\textit{pause\_cmd} \land \neg(\textit{play\_cmd} \lor \textit{ctrl}) \leftrightarrow {}
\\
&\qquad \neg(\textit{play\_cmd} \land \neg\textit{ctrl} \leftrightarrow \textit{ctrl} \land \neg\textit{play\_cmd}) \land \neg\textit{pause\_cmd})\bigr)
\\
&\quad\land\;
    G\bigl(\textit{leave\_app} \rightarrow \neg\textit{play\_button} \land \neg\textit{pause\_button}\bigr) \land G\bigl(\neg(\textit{play\_button} \land \textit{pause\_button})\bigr)
\\
&\quad\land\;
    G\bigl(\neg(\textit{leave\_app} \land \textit{resume\_app})\bigr) \land G\bigl(\textit{play\_cmd} \rightarrow X(\textit{music\_playing}\; W\; \textit{pause\_cmd})\bigr)
\\
&\quad\land\;
    G\bigl(\textit{pause\_cmd} \rightarrow X(\neg\textit{music\_playing}\; W\; \textit{play\_cmd})\bigr)
\Bigr)
\\
&\rightarrow 
\Bigl(
    G\bigl(\textit{play\_button} \rightarrow \textit{play\_cmd}\bigr) \land G\bigl(\textit{pause\_button} \rightarrow \textit{pause\_cmd}\bigr)
\\
&\quad\land\;
    G\bigl(\textit{pause\_cmd} \rightarrow \textit{leave\_app} \lor \textit{pause\_button}\bigr) \land G\bigl(\textit{play\_cmd} \rightarrow \neg\textit{leave\_app}\bigr)
\\
&\quad\land\;
    G\bigl(\textit{pause\_cmd} \land \textit{pause\_button} \rightarrow \neg\textit{play\_cmd}\; W\; \textit{play\_button}\bigr)
\\
&\quad\land\;
    G\Bigl(\textit{music\_playing} \land \textit{leave\_app} \rightarrow \bigl(\neg(\textit{pause\_cmd} \land {}
\\
&\qquad (\neg(\neg\textit{pause\_button} \rightarrow \textit{play\_cmd})\; W\; (\neg\textit{pause\_button} \rightarrow \textit{play\_cmd}) \land \neg\textit{leave\_app}))\; W\;
\\
&\qquad \textit{pause\_cmd} \land (\neg(\neg\textit{pause\_button} \rightarrow \textit{play\_cmd})\; W\; {}
\\
&\qquad\quad (\neg\textit{pause\_button} \rightarrow \textit{play\_cmd}) \land \neg\textit{leave\_app}) \land \textit{leave\_app}\bigr)
\Bigr)
\Bigr)
\end{aligned}
\]
with 
\[
I = \{\textit{leave\_app}, \textit{music\_playing}, \textit{pause\_button}, \textit{play\_button}, \textit{resume\_app}\},\text{ } O = \{\textit{ctrl}, \textit{pause\_cmd}, \textit{play\_cmd}\}
\]
Given \(\varphi\), consider the trace  
\[
\pi = \{\text{pause\_cmd},\; \text{leave\_app},\; \text{pause\_button};\;
\text{play\_button},\; \text{play\_cmd},\; \text{leave\_app},\; \text{play\_button}\}
\footnote{Only positive atomic propositions are denoted.}
\]
and let  
\[
E = \text{XX play\_cmd}.
\]
\end{example}
The instantiation of temporal causality as seen in Figure~\ref{fig:music-player-hoa} for the music player \(T\)'s behavior in \(\pi\) demonstrates that the set of inputs required to trigger \(\text{play\_cmd}\) at \(t_{3}\) is much greater than the \(\text{play\_button}\) input at \(t_{2}\), as implied by \(\varphi\). 
Understanding these unintuitive relationships helps develop deeper temporal reasoning about various reactive systems and enables a more efficient way to generate temporal reasoning for LLMs in training and chain-of-thought (CoT) prompting.

\FloatBarrier
\section{Qualitative Study: Narrative Enrichment with Counterfactual Access}
\label{app:qualitative-enrichment}

A natural question is whether translating the formal Mealy automaton specification into a rich natural-language narrative helps or hinders LLM causal reasoning.
We present Claude Sonnet~4.6 with a causal-reasoning task on the arbiter automaton shown in Figure~\ref{fig:qual-automaton}, reformulated as an immersive reactor-meltdown scenario, giving the model all counterfactual information needed to solve the task, but expressed entirely in narrative terms.

\subsection{Task Setup}
\label{app:qual-setup}

The test sample is an 8-step causal reasoning task (\texttt{sample\_id: 8ccd0656}) on the automaton in Figure~\ref{fig:qual-automaton}.
Boolean inputs are mapped to physical actions (\texttt{try\_0}~$\to$~``engaging the Primary Coolant Loop''; \texttt{try\_1}~$\to$~``engaging the Secondary Thermal Sink'') and outputs to diagnostic postures.
The gold answer identifies \textbf{7 of 8 steps as ``no constraints''} (any input produces the same output) and \textbf{one step (Step~6) where \texttt{try\_1} is causally necessary}---flipping it changes the next state, which alters downstream outputs.

\begin{figure}[h]
\centering
\begin{tikzpicture}[
    ->,>=Stealth,
    shorten >=1pt,
    node distance=2.2cm and 2.8cm,
    every state/.style={minimum size=22pt, inner sep=1pt, font=\footnotesize},
    every edge/.style={draw, font=\scriptsize},
    scale=0.9, transform shape,
]
  \node[state, initial, initial text={}] (s0) {$S_0$};
  \node[state, above right=1.2cm and 2.2cm of s0] (s1) {$S_1$};
  \node[state, right=2.2cm of s0] (s2) {$S_2$};
  \node[state, below=2.2cm of s1] (s3) {$S_3$};
  \node[state, below=2.2cm of s2] (s4) {$S_4$};
  \node[state, right=2.8cm of s3] (s5) {$S_5$};
  \node[state, right=2.8cm of s4] (s6) {$S_6$};

  \path (s0) edge[above left] node[align=center]{$\overline{t_0}\,\overline{t_1}$} (s1)
        (s0) edge[above] node{$\overline{t_0}\,t_1$} (s2)
        (s0) edge[below left] node[align=center]{$t_0\,\overline{t_1}$} (s3)
        (s0) edge[below] node[swap]{$t_0\,t_1$} (s4);

  \path (s1) edge[above] node{$\overline{t_0}$} (s5)
        (s1) edge[above right] node{$t_0$} (s6)
        (s2) edge[below left, bend right=10] node[swap]{$\overline{t_0}$} (s5)
        (s2) edge[below right, bend left=10] node{$t_0$} (s6)
        (s3) edge[above] node{$\overline{t_0}$} (s5)
        (s3) edge[below left, bend right=15] node[swap, pos=0.4]{$t_0$} (s6)
        (s4) edge[below left, bend right=10] node[swap]{$\overline{t_0}$} (s5)
        (s4) edge[below] node{$t_0$} (s6);

  \path (s5) edge[bend left=15] node[right, pos=0.4]{$\overline{t_1}$} (s3)
        (s5) edge[bend left=15] node[left, pos=0.4]{$t_1$} (s4)
        (s6) edge[bend right=15] node[left, pos=0.4]{$\overline{t_1}$} (s3)
        (s6) edge[bend right=15] node[right, pos=0.4]{$t_1$} (s4);

  \node[draw=none, text=black!50, font=\tiny, align=center] at ($(s0)+(0, 1.2)$) {All outputs from $S_0$:\\$\overline{g_0}\,\overline{g_1}$};
\end{tikzpicture}
\caption{Arbiter automaton used in the qualitative study (7 states, 16 transitions). Inputs $t_0$, $t_1$ correspond to \texttt{try\_0}, \texttt{try\_1}. A key structural property: from any given state, \emph{all} outgoing transitions produce the same output---making most inputs causally unnecessary. The exception arises at Step~6 (from $S_5$), where flipping $t_1$ redirects to $S_3$ instead of $S_4$, changing the \emph{downstream} output at Step~7.}
\label{fig:qual-automaton}
\end{figure}

\subsection{Enriched Narrative with Observation Archive}
\label{app:qual-enriched}

The prompt wraps the automaton in a reactor-meltdown scenario and includes a \emph{Historical Observation Archive}: a complete enumeration of all transitions, expressed as prior operational experience.
This gives the model full counterfactual access in narrative form. The complete prompt is shown below, followed by the model's response.

\begin{lstlisting}[caption={Complete enriched prompt sent to Sonnet~4.6}]
The Icarus Protocol: Reactor Meltdown Logs

The Icarus Station orbital mining rig was never meant to sustain this
much thermal stress. Deep in the belly of the facility, the master
reactor's logic mainframe is spiraling out of control. The engineering
crew has only two levers: they can either engage or bypass the Primary
Coolant Loop, and they can engage or bypass the Secondary Thermal Sink.

Every action forces the reactor's automated safety matrix to
re-evaluate survivability across four parameters: Core Stability,
Plasma Containment, Automated Telemetry, and Manual Logs.

### Historical Observation Archive

Before the crisis, the station logged every known reactor response
across all possible configurations:

From Phase S0 (Initial Cold Start):
- Bypass BOTH: Sub-system Dormancy, all feeds dark. (S0 -> S1)
- Bypass Coolant, engage Sink: Total System Shutdown. (S0 -> S2)
- Engage Coolant, bypass Sink: Isolated Core Ignition. (S0 -> S3)
- Engage BOTH: Synchronized Thermal Purge. (S0 -> S4)

From Phase S1 (Sub-system Dormancy):
- Bypass Coolant: Abyssal Coolant Purge. (S1 -> S5)
- Engage Coolant: Abyssal Coolant Purge. (S1 -> S6)

From Phase S2 (Total System Shutdown):
- Bypass Coolant: Surface-Directed Venting. (S2 -> S5)
- Engage Coolant: Plasma Containment restored, core unstable. (S2 -> S6)

From Phase S3 (Isolated Core Ignition):
- Bypass Coolant: Surface-Directed Venting. (S3 -> S5)
- Engage Coolant: Surface-Monitored Ignition. (S3 -> S6)

From Phase S4 (Synchronized Thermal Purge):
- Bypass Coolant: Full Ack. Venting, dual logs, containment. (S4 -> S5)
- Engage Coolant: Full Ack. Venting, dual logs, containment. (S4 -> S6)

From Phase S5:
- Bypass Thermal Sink: Isolated Core Ignition. (S5 -> S3)
- Engage Thermal Sink: Synchronized Thermal Purge. (S5 -> S4)

From Phase S6:
- Bypass Thermal Sink: Isolated Core Ignition with ack. (S6 -> S3)
- Engage Thermal Sink: Synch. Thermal Purge with ack. (S6 -> S4)

### The Operational Log

Day 1: Bypassed coolant loop, engaged thermal sink.
       Total system shutdown. (S0 -> S2)
Day 2: Engaged coolant loop. Plasma containment restored,
       core unstable. (S2 -> S6)
Day 3: Engaged thermal sink. Core stabilized, containment
       collapsed. (S6 -> S4)
Day 4: Re-engaged coolant loop. Dual logs, containment
       restored, core crashed. (S4 -> S6)
Day 5: Engaged thermal sink. Core stability returned,
       containment collapsed. (S6 -> S4)
Day 6: Bypassed coolant loop. Dual logs, containment sealed,
       core unstable. (S4 -> S5)
Day 7: Engaged thermal sink. Telemetry blind, core stabilized,
       containment collapsed. (S5 -> S4)
Day 8: Bypassed coolant loop. Containment held, core offline.
       (S4 -> S5)

### Task
For each day, determine which crew actions were causally necessary.
Use the Historical Observation Archive: if the crew had done the
opposite, would the reactor's response have changed?
\end{lstlisting}

The model's response excerpts:

\begin{lstlisting}[caption={Enriched Format with Observation Archive: Sonnet~4.6 response (excerpts)}]
Day 1 (Step 0): "Both actions causally necessary"
  Counterfactual: engaging coolant -> S0->S4; bypassing sink -> S0->S1
  "Either alternative would have produced a different outcome."

Day 6 (Step 5): "INEVITABLE OUTCOME DETECTED"
  "From S4, both coolant loop options produce functionally identical
   results... The crew's action was functionally irrelevant."

Day 7 (Step 6): "Thermal sink engagement was causally necessary"
  Counterfactual: bypassing thermal sink -> S5->S3
  "The bypass would have produced a different outcome."

Day 8 (Step 7): "INEVITABLE OUTCOME DETECTED"
  "From S4, the reactor's response was functionally predetermined
   regardless of coolant loop action."
\end{lstlisting}

\subsection{Analysis}
\label{app:qual-analysis}

\begin{table}[h]
\centering
\small
\begin{tabular}{lcccl}
\toprule
\textbf{Step} & \textbf{Gold} & \textbf{Model} & & \textbf{Failure Mode} \\
\midrule
0 & no constraints & \texttt{try\_0}, \texttt{try\_1} & \ding{55} & State $\neq$ output causality \\
1 & no constraints & \texttt{try\_0} & \ding{55} & State $\neq$ output causality \\
2 & no constraints & \texttt{try\_1} & \ding{55} & State $\neq$ output causality \\
3 & no constraints & \texttt{try\_0} & \ding{55} & State $\neq$ output causality \\
4 & no constraints & \texttt{try\_1} & \ding{55} & State $\neq$ output causality \\
5 & no constraints & no constraints & \ding{51} & --- \\
6 & \texttt{try\_1} & \texttt{try\_1} & \ding{51} & --- \\
7 & no constraints & no constraints & \ding{51} & --- \\
\bottomrule
\end{tabular}
\caption{Per-step correctness of Claude Sonnet~4.6 on the enriched narrative with full counterfactual access via the observation archive (3/8 correct). The model correctly identifies the \emph{one} causally necessary input (Step~6) and two ``no constraints'' steps, but produces 5 false positives by conflating state-level and output-level causality.}
\label{tab:qualitative-enrichment}
\end{table}

With the observation archive, the model gains the ability to perform genuine counterfactual checks and \textbf{correctly identifies Step~6}---the single step where an input is causally necessary---which it misses entirely in the formal format.
However, it introduces \textbf{5 false positives} by conflating two notions of causality:

\begin{enumerate}[leftmargin=*, nosep]
\item \textbf{State causality}: flipping an input changes the \emph{next state} (e.g., $S_0 \to S_1$ vs.\ $S_0 \to S_4$).
\item \textbf{Output causality}: flipping an input changes the \emph{produced output}.
\end{enumerate}

In the formal format, output equivalence is transparent---the model can see that both transitions from $S_0$ produce the identical string \texttt{!g\_0\&!g\_1}.
In the narrative format, these same outputs receive \emph{different narrative descriptions} (``Sub-system Dormancy'' vs.\ ``Synchronized Thermal Purge''), making them appear causally distinct even when the underlying outputs are identical.

The observation archive thus creates a revealing tradeoff.
It provides enough structure for the model to correctly detect the one genuine causal dependency (Step~6: different next states lead to different \emph{downstream} outputs), but the narrative framing obscures output equivalence at every other step, producing a net decrease in accuracy (3/8 vs.\ 7/8 in the formal format).
This suggests that the difficulty of MIN is not merely about access to counterfactual information, but about the \emph{representation} in which that information is expressed: formal notation makes output equivalence trivially verifiable, while natural language systematically obscures it.

\FloatBarrier
\section{Truncated Visibility: Qualitative Examples}
\label{app:truncated-output-samples}

The following excerpts show the same SIM task (arbiter\_with\_buffer\_prio\_param\_n2, path length 7) evaluated under three visibility conditions. At 100\%, the model achieves 7/7 correct steps; at 50\%, 1/7; at 25\%, 0/7. Model output is in black; \textcolor{aiblue}{blue annotations} are ours.

\textbf{100\% visibility (7/7 correct).} The model sees all 4 transitions from state 0, systematically checks each transition condition, and correctly identifies the matching transition:
\begin{lstlisting}
At step 0, the machine is in state 0.
Input: 'try_0&try_1'

The transitions from state 0 are:
  - Transition to state 1 produces '!g_0&!g_1'
    AND-guard fails: try_1=true FAILS '!try_1' (needs false).
  - Transition to state 2 produces '!g_0&!g_1'
    AND-guard fails: try_1=true FAILS '!try_1' (needs false).
  - Transition to state 3 produces '!g_0&!g_1'
    AND-guard fails: try_1=true FAILS '!try_1' (needs false).
  - Transition to state 4 produces '!g_0&!g_1'
    AND-guard evaluates to true: all 2 conditions met.

Therefore, output='!g_0&!g_1' and we move to state 4.
(*@\textcolor{aiblue}{\rmfamily CORRECT: Correct state (4) and output. All 4 guards checked systematically.}@*)

At step 1, the machine is in state 4. Input: 'try_0'
  - Transition to state 5: Guard '!try_0' fails.
  - Transition to state 6: Guard 'try_0' evaluates to true.
The machine outputs 'ack_0&ack_1&!g_0&g_1' and transitions to state 6.
(*@\textcolor{aiblue}{\rmfamily CORRECT: Correct. Model continues accurately through all 7 steps.}@*)
\end{lstlisting}

\textbf{50\% visibility (1/7 correct).} Only 2 of 4 transitions from state 0 are visible. The model correctly rejects both but must guess the next state:

\begin{lstlisting}
At step 0, the machine is in state 0.
Input: 'try_0&try_1'

The transitions from state 0 are:
  - Transition to state 2 producing '!g_0&!g_1' - NO MATCH
  - Transition to state 3 producing '!g_0&!g_1' - NO MATCH

The machine outputs '!g_0&!g_1' and transitions to state 3.
(*@\textcolor{aiblue}{\rmfamily ERROR: Output correct by luck (all transitions produce !g\_0\&!g\_1).}@*)
(*@\textcolor{aiblue}{\rmfamily ERROR: Wrong state: picked 3 instead of 4 (transition to 4 was hidden).}@*)

At step 1, the machine is in state 3. (*@\textcolor{aiblue}{\rmfamily (ERROR: wrong state)}@*)
Input: 'try_0'
  - Transition to state 5 producing 'ack_0&!ack_1&!g_0&g_1' - NO MATCH
The machine outputs 'ack_0&!ack_1&!g_0&g_1' and transitions to state 5.
(*@\textcolor{aiblue}{\rmfamily ERROR: Wrong output and state. Errors cascade from step 0.}@*)
\end{lstlisting}

\textbf{25\% visibility (0/7 correct).} With 75\% hidden, the model sees wrong transitions with wrong outputs and hallucinates condition satisfaction:

\begin{lstlisting}
At step 0, the machine is in state 0.
Input: 'try_0&try_1'

The transitions from state 0 are:
  - Transition to state 1 produces 'ack_0&!ack_1&!g_0&g_1' - NO MATCH
  - Transition to state 2 produces 'ack_0&!ack_1&g_0&!g_1' - NO MATCH
  - Transition to state 5 produces 'ack_0&!ack_1&!g_0&g_1'
    This transition FIRES because try_0=true satisfies '!try_0'
(*@\textcolor{aiblue}{\rmfamily ERROR: HALLUCINATION: try\_0=true does NOT satisfy !try\_0.}@*)
(*@\textcolor{aiblue}{\rmfamily ERROR: Visible transitions have wrong outputs (ack\_0\&... instead of !g\_0\&!g\_1).}@*)
(*@\textcolor{aiblue}{\rmfamily ERROR: Correct transition (to state 4) is not visible.}@*)

The machine outputs 'ack_0&!ack_1&!g_0&g_1' and transitions to state 5.

At step 1, the machine is in state 5. Input: 'try_0'
  - Transition to state 3: NO MATCH
  - Transition to state 4: NO MATCH
  - Transition to state 5: NO MATCH
  - Transition to state 6: NO MATCH
The machine outputs '!ack_0&g_0&!g_1' and transitions to state 5.
(*@\textcolor{aiblue}{\rmfamily ERROR: All visible transitions rejected but model guesses anyway.}@*)
(*@\textcolor{aiblue}{\rmfamily ERROR: Stuck in wrong state 5, wrong outputs for all remaining steps.}@*)
\end{lstlisting}

The progression shows that the model actively reads the transition table for SIM: at 100\% it checks transition conditions systematically, at 50\% it correctly rejects visible conditions but guesses wrong states, and at 25\% it hallucinates condition satisfaction to force a match. The increasing condition copy rate in MIN (Table~\ref{tab:memorization_test}: 48\% $\to$ 55\%) under reduced visibility confirms the model also reads the transition table for MIN, but uses it for retrieval (copying conditions) rather than reasoning (counterfactual analysis).

\section{Sample Prompts}
\label{app:prompts}

\subsection{SIM: Trace Simulation Prompt}
\label{app:q1-prompt}

The following prompt is used to evaluate trace simulation capability. The model must simulate the automaton step-by-step and report the input read, output produced, and resulting state at each transition.

\begin{lstlisting}[caption={Q1 Evaluation Prompt}]
You are an expert systems analyst reviewing a temporal protocol.

## Q1 Task:
It is your task to use the inputs and outputs in the trace to check what
transition has been taken.

Output the sequence of transitions and states exactly as shown in the question
instructions.

### System Description
SYSTEM:
This is a Mealy automaton with 8 states and 16 transitions.
The automaton can read the following symbols: !try_0, !try_0 & !try_1,
!try_0 & try_1, !try_1, try_0, try_0 & !try_1, try_0 & try_1, try_1.
Initial state(s): 0.

Transition functions:
  - From 0 to 1 on input '!try_0&!try_1', producing output '!g_0&!g_1'.
  - From 0 to 2 on input '!try_0&try_1', producing output '!g_0&!g_1'.
  - From 0 to 3 on input 'try_0&!try_1', producing output '!g_0&!g_1'.
  - From 0 to 4 on input 'try_0&try_1', producing output '!g_0&!g_1'.
  [... additional transitions ...]

### Trace
TRACE:
The automaton processes the following input sequence of length 9:
ack_0&!ack_1&!g_0&!g_1&r_0&r_1&!try_0&try_1;
!ack_0&ack_1&!g_0&g_1&r_0&r_1&!try_0&!try_1;
[... additional steps ...]

### Questions
q1: For each transition step_k in the trace (k = 0,...,T), output the exact
input label read, the output produced, and the resulting next_state in order.

## Output format
{
  "answers": {
    "q1": {
      "step_0": {"input": "<...>", "output": "<...>", "next_state": "<id>"},
      "step_1": {"input": "<...>", "output": "<...>", "next_state": "<id>"},
      ...
    }
  }
}
\end{lstlisting}

\subsection{MIN: Causal Minimality Prompt}
\label{app:q2-prompt}

The following prompt evaluates causal minimality. The model must identify the minimal set of input constraints necessary for each observed output---requiring counterfactual reasoning about which inputs could have been different without changing the outcome.

\begin{lstlisting}[caption={Q2 Evaluation Prompt}]
You are an expert systems analyst reviewing a temporal protocol.

## Q2 Task: Identifying Required Constraints

You are given a discrete state-transition system (automaton), a current state
s_k, and an observed transition at step k that produces a specific output o_k
and next state s_{k+1} under some input valuation.

Your task is to identify the minimal set of input literals that are *causally
required* for this transition to occur.

Formally, for each step_k, output a set of input literals C_k such that:

1. **Sufficiency**:
   There exists at least one complete input valuation that satisfies all
   literals in C_k and causes the automaton to produce exactly output o_k
   and transition to state s_{k+1}.

2. **Necessity**:
   For every literal l in C_k, removing l makes the transition impossible:
   there exists no input valuation satisfying C_k \ {l} that would produce
   both output o_k and next state s_{k+1} from state s_k.

3. **Minimality**:
   C_k contains no redundant literals.

If no input literals are required, output "no constraints".

### System Description
[Same automaton definition as Q1]

### Trace
[Same trace as Q1]

### Questions
q2: For each step_k, list the minimal required input literals (or 'no
constraints') such that removing any listed literal would make it impossible
for the automaton to produce the same output at step_k.

## Output format
{
  "answers": {
    "q2": {
      "step_0": {"required_inputs": ["<lit1>", "<lit2>"] OR "no constraints"},
      "step_1": {"required_inputs": [...] OR "no constraints"},
      ...
    }
  }
}
\end{lstlisting}

\subsection{SFT Training Format}
\label{app:sft-prompt}

For supervised fine-tuning, we use an instruction-output format with detailed chain-of-thought reasoning in the output. The model learns to explicitly trace through the automaton, checking each transition guard.

\begin{lstlisting}[caption={SFT Training Example (Instruction)}]
You are simulating a Mealy automaton. Given the machine definition and input
sequence, predict the output at each step.

### Machine Definition
SYSTEM:
This is a Mealy automaton with 17 states and 72 transitions.
The automaton can read the following symbols: !try_0 & !try_1,
!try_0 & !try_1 & !try_2, !try_0 & !try_1 & try_2, ...
Initial state(s): 0.

Transition functions:
  - From 0 to 1 on input '!try_0&!try_1&!try_2', producing '!g_0&!g_1&!g_2'.
  - From 0 to 2 on input '!try_0&!try_1&try_2', producing '!g_0&!g_1&!g_2'.
  [... additional transitions ...]

### Input Sequence
try_0&!try_1&try_2; !try_0&try_1&!try_2; ...
\end{lstlisting}

\begin{lstlisting}[caption={SFT Training Example (Output with Chain-of-Thought)}]
Working through the automaton, I need to check which transitions fire at each
step.

Starting in state 0 (the initial state).

Now at step 0 in state 0.
Input: 'try_0&!try_1&try_2'
Input values: try_0=true, try_1=false, try_2=true

The transitions from state 0 are:
  - Transition to state 1 with guard '!try_0&!try_1&!try_2':
    NO MATCH: try_0=true FAILS '!try_0' (needs false).
  - Transition to state 2 with guard '!try_0&!try_1&try_2':
    NO MATCH: try_0=true FAILS '!try_0' (needs false).
  - Transition to state 6 with guard 'try_0&!try_1&try_2':
    MATCH: try_0=true satisfies 'try_0'; try_1=false satisfies '!try_1';
    try_2=true satisfies 'try_2'.

Taking transition to state 6, output: '!g_0&!g_1&!g_2'

[... continues for each step ...]

Final answer:
{
  "step_0": {"input": "try_0&!try_1&try_2", "output": "!g_0&!g_1&!g_2",
             "next_state": "6"},
  ...
}
\end{lstlisting}

The chain-of-thought format explicitly shows the guard-checking process, teaching the model to systematically evaluate each transition condition rather than pattern-matching on surface features.

\subsection{SFT Training Format: Causal Reasoning}
\label{app:sft-causal}

For causal reasoning, the SFT output demonstrates counterfactual analysis, checking what would happen if each input literal were flipped.

\begin{lstlisting}[caption={Causal Reasoning SFT Example (Instruction)}]
You are performing causal analysis on a Mealy automaton execution. For each
step, determine which input literals are causally necessary - meaning if you
flipped that literal, the output would change.

### Machine Definition
SYSTEM:
This is a Mealy automaton with 8 states and 16 transitions.
[... automaton definition ...]

### Execution Trace
Step 0: state=0, input='!try_0&!try_1', output='!g_0&!g_1', next_state=1
Step 1: state=1, input='!try_0', output='!ack_0&!ack_1&!g_0&g_1', next_state=5
Step 2: state=5, input='try_1', output='!ack_0&g_0&!g_1', next_state=4
[...]
\end{lstlisting}

\begin{lstlisting}[caption={Causal Reasoning SFT Example (Output with Counterfactual Analysis)}]
I need to determine which input literals are causally necessary for the
observed outputs using counterfactual reasoning.

==================================================
Step 0: state=0, input='!try_0&!try_1', output='!g_0&!g_1'
==================================================

Available transitions from state 0:
  - To state 1: guard='!try_0&!try_1', output='!g_0&!g_1' [TAKEN]
  - To state 2: guard='!try_0&try_1', output='!g_0&!g_1'
  - To state 3: guard='try_0&!try_1', output='!g_0&!g_1'
  - To state 4: guard='try_0&try_1', output='!g_0&!g_1'

Counterfactual Analysis:
Current input: !try_0&!try_1
Question: Which input literals are ESSENTIAL for getting output '!g_0&!g_1'?

Analysis: All transitions from state 0 produce compatible results.
  - Checking each alternative transition:
    - Transition to 2 also produces '!g_0&!g_1' - output unchanged
    - Transition to 3 also produces '!g_0&!g_1' - output unchanged
    - Transition to 4 also produces '!g_0&!g_1' - output unchanged

Conclusion: No specific input literals are causally necessary.
  Any valid input from this state yields output '!g_0&!g_1'.

Required inputs: ["no constraints"]

==================================================
Step 1: state=1, input='!try_0', output='!ack_0&!ack_1&!g_0&g_1'
==================================================

Available transitions from state 1:
  - To state 5: guard='!try_0', output='!ack_0&!ack_1&!g_0&g_1' [TAKEN]
  - To state 6: guard='try_0', output='!ack_0&!ack_1&!g_0&g_1'

Counterfactual Analysis:
Question: Which input literals are ESSENTIAL for this output?

Analysis: Both transitions produce the same output '!ack_0&!ack_1&!g_0&g_1'.
  - If we flip !try_0 to try_0: transition to state 6, same output

Conclusion: No specific input literals are causally necessary.

Required inputs: ["no constraints"]

[... additional steps with similar analysis ...]

Final answer:
{
  "step_0": {"required_inputs": "no constraints"},
  "step_1": {"required_inputs": "no constraints"},
  "step_7": {"required_inputs": ["try_0"]},  // Only step with constraints
  ...
}
\end{lstlisting}

The causal reasoning chain-of-thought explicitly enumerates alternative transitions and checks whether each produces the same output, teaching the model to reason about \emph{necessity} rather than merely \emph{sufficiency}.
\newpage
\end{document}